\documentclass{CVM}
\usepackage{multicol}

\CVMsetup{
type      = {Research Article},
doi       = {CVM.XXXX},
title     = {IGRecon4D: Dynamic-Object-Aware Feed-Forward Reconstruction of Driving Scenes},
author    = {Yifei Tong$^{1}$, Ziyang Huang$^{1}$, Yizhao Han$^{1}$, Xiaoxiao Long$^{1}$\cor{}, Hao Zhu$^{1}$\cor{} },
runauthor = {Y. Tong et al.},
abstract  = {
  Feed-forward 4D driving-scene reconstruction enables efficient modeling of dynamic autonomous-driving scenes, but current methods suffer from two complementary failure modes: semantic leakage of traffic participants into the static representation and geometric drift of Gaussian locations across viewpoints. We propose \textbf{IGRecon4D}, a feed-forward dynamic scene reconstruction framework that addresses these failures with training-time instance-prior distillation and LiDAR-anchored Gaussian positioning. Specifically, an instance-level semantic branch provides a potentially dynamic participant prior, which is distilled into dynamic-mask learning through the Instance-Dynamic Consistency Constraint to encourage high-recall localization of traffic-participant regions prone to dynamic-static leakage. In parallel, sparse LiDAR measurements guide dense depth estimation, and the resulting depths are back-projected to anchor Gaussian positions for improved multi-view geometric consistency. Experiments on the Waymo Open Dataset and zero-shot evaluation on nuScenes demonstrate improved full-image reconstruction on the main benchmark and stronger traffic-participant-region reconstruction under cross-dataset evaluation.
},
keywords  = {feed-forward reconstruction; dynamic reconstruction; instance-level semantic learning; autonomous driving},
copyright = {The Author(s)},
}

\begin{document}

\maketitle

\enlargethispage{-3pt}
\begin{figure}[b] \vskip -4mm
\small\renewcommand\arraystretch{1.3}
\begin{tabular}{p{80.5mm}} \toprule\\ \end{tabular}
\vskip -4.5mm \noindent \setlength{\tabcolsep}{1pt}
\begin{tabular}{p{3.5mm}p{80mm}}
$1\quad $ & Department of Intelligent Science and Technology,
Nanjing University, Suzhou 215163, China. E-mail: yifeitong754@outlook.com, hziyang888@gmail.com, 522025710020@smail.nju.edu.cn, xxlong@nju.edu.cn, zh@nju.edu.cn.\\
&\hspace{-5mm} Manuscript received: ; accepted: \vspace{-2mm}
\end{tabular} \vspace {-3mm}
\end{figure}

\section{Introduction}

Dynamic scene reconstruction aims to recover time-varying 3D scenes from visual observations, which is essential for autonomous driving, simulation, and embodied perception. Early implicit methods such as NeRF-based dynamic reconstruction~\cite{mildenhall2020nerf,pumarola2020d,li2020neural} model spatiotemporal variations through neural fields, while explicit methods such as 3D Gaussian Splatting~\cite{kerbl3Dgaussians} and its dynamic extensions~\cite{Wu_2024_CVPR} improve rendering efficiency with Gaussian primitives. However, these methods typically rely on costly per-scene optimization, limiting their scalability to long driving sequences and large-scale datasets.

To reduce per-scene optimization, feed-forward geometric estimation methods have emerged as an efficient alternative. Unlike traditional SfM pipelines such as COLMAP~\cite{schoenberger2016sfm}, these methods directly predict geometric information from one or multiple input views through a single network forward pass. Representative methods such as DUSt3R~\cite{dust3r_cvpr24}, VGGT~\cite{wang2025vggt}, and $\pi^3$~\cite{wang2025pi} have shown strong geometric prediction ability, providing a foundation for recent feed-forward reconstruction methods.

Building on these advances in feed-forward geometric estimation, recent methods further predict Gaussian primitives in a feed-forward manner. In recent years, large reconstruction models (LRMs)~\cite{hong2023lrm} based on the Transformer architecture have demonstrated remarkable progress in 3D reconstruction, particularly multi-view reconstruction methods built upon the Vision Transformer (ViT)~\cite{dosovitskiy2020vit}. Furthermore, several studies have extended feed-forward reconstruction to Gaussian-based scene representations, including Flash3D~\cite{szymanowicz2024flash3d}, MVSplat~\cite{chen2024mvsplat}, NoPoSplat~\cite{ye2024noposplat}, AnySplat~\cite{jiang2025anysplat}, and DepthSplat~\cite{xu2024depthsplat}. These methods achieve efficient 3D reconstruction and high-quality rendering but are primarily designed for static scenes.

For dynamic scenes, recent efforts have begun to explore feed-forward 4D reconstruction. L4GM\cite{ren2024l4gm} introduced the first feed-forward framework for 4D Gaussian reconstruction, but it mainly focuses on object-level modeling and does not readily generalize to complex scene-level reconstruction. Subsequently, STORM\cite{yang2025storm} proposed a feed-forward framework for dynamic scene reconstruction, significantly improving reconstruction efficiency. DGGT\cite{chenfeedforward} further extends feed-forward reconstruction to multi-view autonomous driving scenarios. Nevertheless, existing feed-forward dynamic reconstruction methods still face difficulties in complex scene-level autonomous driving scenarios, where accurate geometry, multi-view consistency, and reliable localization of traffic-participant regions are all required.

Beyond geometry and appearance reconstruction, several feed-forward reconstruction methods also predict additional scene modalities such as instance segmentation. IGGT~\cite{li2025iggt} and EPS3D~\cite{zhu2026eps3dendtoendfeedforward3d} predict instance-level semantics through feature clustering, while SegVGGT~\cite{qu2026segvggt} further improves segmentation quality with a query-based formulation. However, these methods rarely exploit instance-level semantic cues to improve reconstruction itself. We use such cues as a potentially dynamic participant prior for high-recall localization of traffic-participant regions that are prone to dynamic-static leakage.

Building upon these observations, we propose \textbf{IGRecon4D}, a feed-forward framework for dynamic scene reconstruction that jointly addresses Gaussian prediction and traffic-participant-region localization. Our method employs a ViT~\cite{dosovitskiy2020vit} encoder followed by a series of alternating attention layers to extract hierarchical scene features, which are shared by both the reconstruction branch and the instance-level semantic branch.

For dynamic scene reconstruction, a set of prediction heads is used to estimate Gaussian parameters and dynamic masks. To improve the geometric accuracy of the reconstructed scene, we further leverage LiDAR priors to predict more accurate depth, providing more reliable Gaussian positions. The predicted dynamic masks allocate traffic-participant regions that may cause dynamic-static leakage to per-frame Gaussians, while the remaining regions are accumulated into the static Gaussian set. During rendering, each frame is synthesized by jointly rendering the static Gaussians and the dynamic Gaussians corresponding to the current timestamp.

To improve traffic-participant-region reconstruction, we introduce an instance-level semantic branch that provides a potentially dynamic participant prior for dynamic Gaussian modeling. Specifically, the learned instance-level representations supervise dynamic mask prediction through an Instance-Dynamic Consistency Constraint, encouraging high-recall localization of traffic-participant regions that are prone to dynamic-static leakage.

Extensive experiments on the Waymo Open Dataset demonstrate that IGRecon4D improves overall reconstruction quality, traffic-participant-region reconstruction, and novel-view consistency over representative feed-forward reconstruction baselines.

In summary, our main contributions are as follows:

1. We propose \textbf{IGRecon4D}, a feed-forward framework that unifies dynamic 4D Gaussian reconstruction and instance-level semantic learning for autonomous driving scenes.

2. We introduce an Instance-Dynamic Consistency Constraint that exploits instance-level semantic cues as a potentially dynamic participant prior to supervise dynamic mask prediction, leading to improved high-recall localization and reconstruction of traffic-participant regions prone to dynamic-static leakage.

3. We incorporate LiDAR-guided depth estimation to derive reliable Gaussian positions, enabling LiDAR-anchored Gaussian positioning and improving geometric accuracy and multi-view consistency as validated by novel-view synthesis and ablation studies.

4. Extensive experiments on Waymo and zero-shot nuScenes evaluation demonstrate improved full-image reconstruction on the main benchmark and stronger cross-dataset traffic-participant-region reconstruction.

\section{Related Work}

\subsection{Optimization-based Reconstruction}

Dynamic 3D scene reconstruction aims to recover the geometry, appearance, and temporal motion of a scene from monocular or multi-view videos. Compared with static reconstruction, this task must additionally handle moving objects, occlusions, and temporal consistency across frames. Early RGB-D reconstruction methods exploit depth measurements for non-rigid geometry recovery\cite{bozic2020deepdeform}, but their performance depends heavily on sensor quality and often degrades under occlusion, rapid motion, or noisy depth observations.

Neural scene representations have significantly advanced dynamic reconstruction~\cite{10903670}. NeRF-based methods~\cite{mildenhall2020nerf} commonly model dynamics by introducing time-dependent deformation fields or canonical-space mappings, as in D-NeRF~\cite{pumarola2020d}, NeRFies~\cite{park2021nerfiesdeformableneuralradiance}, and Neural 3D Video~\cite{li2022neural3dvideosynthesis}. More recently, dynamic Gaussian-based methods build upon 3D Gaussian Splatting (3DGS)~\cite{kerbl3Dgaussians} by introducing temporal parameters or motion-aware Gaussian representations~\cite{yang2023gs4d,Wang_2025_CVPR,duan:2024:4drotorgs}. These representations improve rendering efficiency and dynamic-scene fidelity, but they still typically rely on per-scene optimization, which limits their scalability to large autonomous-driving datasets.

To improve geometry and temporal consistency under sparse observations~\cite{11724652}, recent dynamic reconstruction systems incorporate additional visual priors such as depth, optical flow, tracking, and semantic or instance cues~\cite{Wu_2024_CVPR,som2024,lei2024mosca,teed2020raftrecurrentallpairsfield,chen2025prodyg,peng2025gpsslam}. These priors help resolve ambiguities in dynamic regions, but they are often integrated through multi-stage pipelines, for example, sequentially estimating depth or flow before reconstruction, resulting in additional system complexity and computational cost. This motivates reconstruction frameworks that can combine efficient feed-forward inference with reliable geometric cues and high-recall localization of traffic-participant regions.

\subsection{Feed-forward Neural Reconstruction}

Feed-forward 3D reconstruction aims to directly predict scene representations, such as NeRFs~\cite{mildenhall2020nerf} or 3D Gaussian primitives~\cite{kerbl3Dgaussians}, through a single forward pass of a neural network, thereby eliminating the costly per-scene optimization required by conventional reconstruction methods. Compared with optimization-based approaches, feed-forward methods offer significantly higher inference efficiency and stronger cross-scene generalization, making them an increasingly important research direction in recent years.

An important precursor to feed-forward 3D reconstruction is feed-forward geometric estimation~\cite{11366109}, which aims to recover scene geometry with a single neural network forward pass. Unlike traditional per-scene SfM pipelines such as COLMAP~\cite{schoenberger2016sfm}, these methods take one or multiple images as input and directly predict geometric information such as depth maps, point maps, camera poses, or tracks. DPT~\cite{ranftl2021visiontransformersdenseprediction} estimates dense depth from a single image with a ViT-based architecture, while DUSt3R~\cite{dust3r_cvpr24} extends feed-forward geometric reconstruction to two input views and enforces two-view geometric consistency. VGGT~\cite{wang2025vggt} further supports reconstruction from more input views and jointly estimates depth, camera poses, and tracks, whereas $\pi^3$~\cite{wang2025pi} decouples reconstruction from the ordering of input views, improving the robustness of feed-forward geometric prediction. These methods provide effective geometric priors for 3D reconstruction and motivate subsequent feed-forward methods to move beyond geometry prediction toward renderable 3D representation reconstruction.

Building on this direction, early feed-forward reconstruction methods mainly focus on object-level reconstruction. Representative approaches such as LRM~\cite{hong2023lrm} and LGM~\cite{ren2024l4gm} learn ViT-based reconstruction priors and can efficiently recover high-quality 3D representations from only a few input views. By learning rich geometric priors from large-scale datasets, these methods substantially improve reconstruction speed while maintaining competitive reconstruction quality. More recently, feed-forward reconstruction has been extended from object-level to scene-level modeling. Methods such as Flash3D~\cite{szymanowicz2024flash3d}, MVSplat~\cite{chen2024mvsplat}, NoPoSplat~\cite{ye2024noposplat}, DepthSplat~\cite{xu2024depthsplat}, and AnySplat~\cite{jiang2025anysplat} leverage feed-forward prediction of 3D Gaussian representations to achieve an effective balance between reconstruction quality and computational efficiency. Unlike optimization-based Gaussian reconstruction methods, these approaches directly infer Gaussian parameters from input observations, enabling efficient scene reconstruction and rendering. Nevertheless, they are primarily designed for static scenes and cannot be readily generalized to dynamic environments.

To address dynamic scene reconstruction, recent studies have begun to investigate feed-forward 4D scene modeling. L4GM~\cite{ren2024l4gm} introduced the first feed-forward framework for 4D Gaussian reconstruction, demonstrating the feasibility of directly predicting dynamic Gaussian representations. However, its design mainly targets object-level reconstruction and does not readily generalize to complex scene-level environments. More recently, DrivingForward~\cite{tian2025drivingforward}, STORM~\cite{yang2025storm}, NoPo4D~\cite{balice2026poseproblem4dfeedforward}, 4DGT~\cite{xu20254dgt}, DrivingRecon~\cite{Lu2024DrivingRecon}, DrivingScene~\cite{11463217}, DynamicVGGT~\cite{He_2026_CVPR}, and StreetForward~\cite{yu2026streetforward} have extended the feed-forward paradigm to dynamic scene reconstruction, significantly alleviating the computational burden of per-scene optimization.

DGGT~\cite{chenfeedforward} is closely related to our setting, as it supports feed-forward dynamic reconstruction from multi-view autonomous-driving videos. Nevertheless, existing feed-forward dynamic reconstruction methods still face difficulties in complex driving scenes, where accurate geometry, multi-view consistency, and traffic-participant-region localization are all required. In particular, insufficient localization of traffic-participant regions can cause regions prone to dynamic-static leakage to be modeled as static content, while inaccurate geometry weakens cross-view consistency. Overall, dynamic-static separation remains a central challenge for dynamic reconstruction methods, since the ability to disentangle and model moving objects and the static background is critical to the final reconstruction quality.

\subsection{Joint Geometric Reconstruction and Multimodal Prediction}

Some feed-forward reconstruction methods go beyond scene reconstruction by predicting additional modalities, such as instance segmentation, while inferring scene geometry and appearance in a single forward pass. Built upon the feed-forward architecture of VGGT~\cite{wang2025vggt}, IGGT~\cite{li2025iggt} and EPS3D~\cite{zhu2026eps3dendtoendfeedforward3d} reconstruct the scene and obtain instance segmentation by predicting instance features and clustering them. To alleviate the segmentation degradation caused by the instability of clustering-based methods, SegVGGT~\cite{qu2026segvggt} adopts a query-based formulation commonly used in instance segmentation and queries multi-level features to further improve segmentation quality. Although these methods predict instance-level information during reconstruction, the interaction between different modalities remains limited. MVGGT~\cite{wu2026mvggt} further takes an additional text prompt as input and localizes the corresponding target in static scenes, while $G^2TAM$~\cite{zhu2026g2tamgeometrygroundedtrack} extends this capability to dynamic scenes. These methods show that feed-forward reconstruction can be extended toward multimodal scene understanding; however, they do not fully exploit auxiliary modalities such as instance segmentation to improve reconstruction itself. For dynamic scene reconstruction, we treat instance-level semantic cues as a potentially dynamic participant prior rather than true motion-state labels. Leveraging this prior for high-recall localization of traffic-participant regions that are prone to dynamic-static leakage can improve traffic-participant-region reconstruction and overall reconstruction quality.

\section{Method}

\begin{figure*}[t]
    \centering
    \includegraphics[width=.96\textwidth]{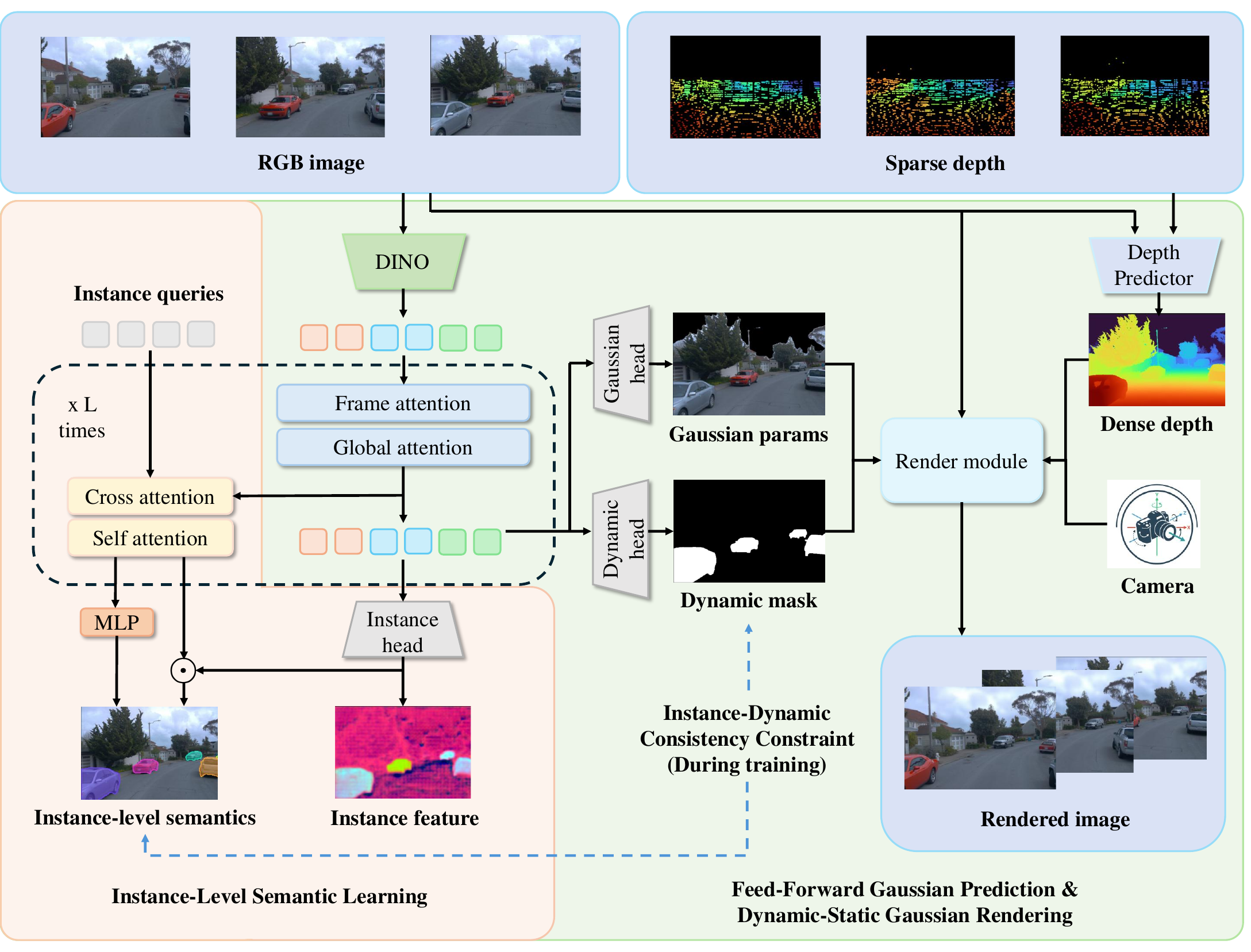}
    \caption{\textbf{Overview of IGRecon4D.} Given multi-camera videos, camera parameters, and sparse LiDAR depth maps, IGRecon4D predicts Gaussian attributes, LiDAR-anchored Gaussian positions, dynamic masks, and instance-level semantic masks. The instance-level semantic branch supplies a potentially dynamic participant prior for dynamic-mask supervision, while the dynamic masks decompose the scene into static and dynamic Gaussians for rendering.}
    \label{fig:pipeline}
\end{figure*}

Given multi-camera videos, camera parameters, and sparse LiDAR depth maps from an autonomous driving scene, IGRecon4D reconstructs a dynamic 4D Gaussian representation in a feed-forward manner. The framework contains two coupled branches: a reconstruction branch that predicts Gaussian attributes, dynamic masks, and Gaussian positions derived from LiDAR-guided depth estimation, and an instance-level semantic branch that learns instance-level cues as a potentially dynamic participant prior for mask supervision. An overview of the proposed framework is illustrated in Fig.~\ref{fig:pipeline}.

The reconstruction branch first extracts hierarchical spatiotemporal features with a ViT backbone and alternating attention layers. It predicts Gaussian attributes and dynamic masks from these features, while dense depth estimated from sparse LiDAR measurements is back-projected to obtain geometrically reliable Gaussian positions. The resulting Gaussians are then decomposed into dynamic and static components and rendered with a differentiable rasterizer.

The instance-level semantic branch shares the reconstruction features and predicts instance masks through learnable queries. During training, these instance predictions provide a potentially dynamic participant prior for dynamic masks through the proposed Instance-Dynamic Consistency Constraint (IDC), encouraging high-recall localization of traffic-participant regions that are prone to dynamic-static leakage. The following subsections describe the dynamic 4D Gaussian representation (Sec.~\ref{sec:scene}), feed-forward Gaussian prediction and LiDAR-guided depth estimation (Sec.~\ref{sec:feedforward}), dynamic-static rendering (Sec.~\ref{sec:render}), instance-level semantic learning (Sec.~\ref{sec:instance}), and the two-stage training objectives (Sec.~\ref{sec:train}).

\subsection{Dynamic 4D Gaussian Representation}
\label{sec:scene}

Dynamic scene reconstruction requires a representation that captures per-frame geometry and appearance while maintaining temporal persistence across frames. We therefore follow DGGT~\cite{chenfeedforward} and represent each frame using a set of pixel-aligned 3D Gaussians. Specifically, for the frame at time step $t$, we denote its Gaussian representation by $G^{t}$, where $G^{t} \in \mathbb{R}^{H \times W \times 15}$. Each pixel $(i,j)$ corresponds to a Gaussian primitive parameterized by its color $c^{t}_{i,j} \in \mathbb{R}^{3}$, position $\mu^{t}_{i,j} \in \mathbb{R}^{3}$, rotation quaternion $r^{t}_{i,j} \in \mathbb{R}^{4}$, scale $s^{t}_{i,j} \in \mathbb{R}^{3}$, opacity $o^{t}_{i,j} \in \mathbb{R}$, and a positive lifespan parameter $\sigma^{t}_{i,j} \in \mathbb{R}^{+}$.

The lifespan parameter controls the temporal existence of each Gaussian primitive by modulating its opacity across different timestamps. Specifically, for a Gaussian defined at time step $t$, its opacity at another time step $t'$ is computed as:

\begin{equation}
    o^{t'} = o^{t} \cdot
    \exp\left(
    -\frac{1}{2}
    \cdot
    \frac{(t'-t)^2}{\sigma^{t}}
    \right).
\end{equation}

A larger lifetime parameter $\sigma$ indicates that the corresponding Gaussian persists for a longer temporal duration, whereas a smaller $\sigma$ results in a shorter lifespan.

\subsection{Feed-Forward Gaussian Prediction}
\label{sec:feedforward}

We first partition the input video into a sequence of image patches, which are then fed into a ViT encoder to extract visual representations. Following DGGT~\cite{chenfeedforward}, we adopt the feature extraction backbone of DINOv2\cite{oquab2024dinov2learningrobustvisual} as our ViT\cite{dosovitskiy2020vit} encoder. The extracted visual features, denoted by $F_{\mathrm{dino}}$, are subsequently passed through a stack of alternating attention layers. Following the design of VGGT\cite{wang2025vggt}, each alternating attention block consists of a frame attention layer and a global attention layer. By progressively aggregating spatial and temporal information, these layers produce a set of hierarchical features, denoted by $F_{\mathrm{attn}}$, which contains the outputs of every frame attention and global attention block.

We then employ a prediction head to estimate the Gaussian parameters except for the position, together with the dynamic mask. The architecture of the prediction head follows the design of the decoder proposed in DPT\cite{ranftl2021visiontransformersdenseprediction}. Let $H_{gs}$ denote the Gaussian prediction head, and let
\[
A^{t}=\{c^{t},r^{t},s^{t},o^{t},\sigma^{t}\}
\]
represent the Gaussian parameters excluding the position. The prediction is formulated as

\begin{equation}
A^{t}=H_{gs}(F_{\mathrm{attn}},F_{\mathrm{dino}}).
\end{equation}

Both $F_{\mathrm{attn}}$ and $F_{\mathrm{dino}}$ are utilized for Gaussian parameter prediction because they provide complementary information. Specifically, $F_{\mathrm{attn}}$ captures high-level semantic and temporal representations through the alternating attention blocks, whereas $F_{\mathrm{dino}}$ preserves richer low-level visual details. By jointly exploiting hierarchical features from different levels, the prediction head is able to estimate the Gaussian parameters more accurately.

Accurate Gaussian positions are critical for multi-view consistency, but directly regressing 3D positions from images is unstable in large-scale driving scenes. To obtain Gaussian positions, we take the input video together with the sparse depth maps
$D^{\mathrm{sparse}}_{t}$ obtained by projecting LiDAR measurements onto the image plane using given camera parameters. Further details of the LiDAR projection are provided in Appendix~\ref{app:lidar_projection}. A pretrained depth estimation network, InfiniDepth\cite{yu2026infinidepth}, is employed to recover dense depth maps
$D^{\mathrm{dense}}_{t}$ from the sparse observations. The dense depth maps are back-projected into Gaussian positions in the rendering module.

In parallel, a dynamic-mask prediction head, denoted by $H_{\mathrm{dyn}}$, is employed to predict a dynamic mask
$M^{t}$, which represents the probability that each pixel belongs to a traffic-participant region that may cause dynamic-static leakage if absorbed into the static representation. The term ``dynamic'' here denotes reconstruction allocation to per-frame Gaussians rather than a ground-truth motion-state label. The prediction is formulated as

\begin{equation}
M^{t}=H_{\mathrm{dyn}}(F_{\mathrm{attn}}).
\end{equation}

The predicted dynamic mask is subsequently used in the Gaussian rendering module to allocate high-prior traffic-participant regions to per-frame Gaussian primitives instead of the global static representation.

\subsection{Dynamic-Static Gaussian Rendering}
\label{sec:render}

\begin{figure*}[t]
    \centering
    \includegraphics[width=\linewidth]{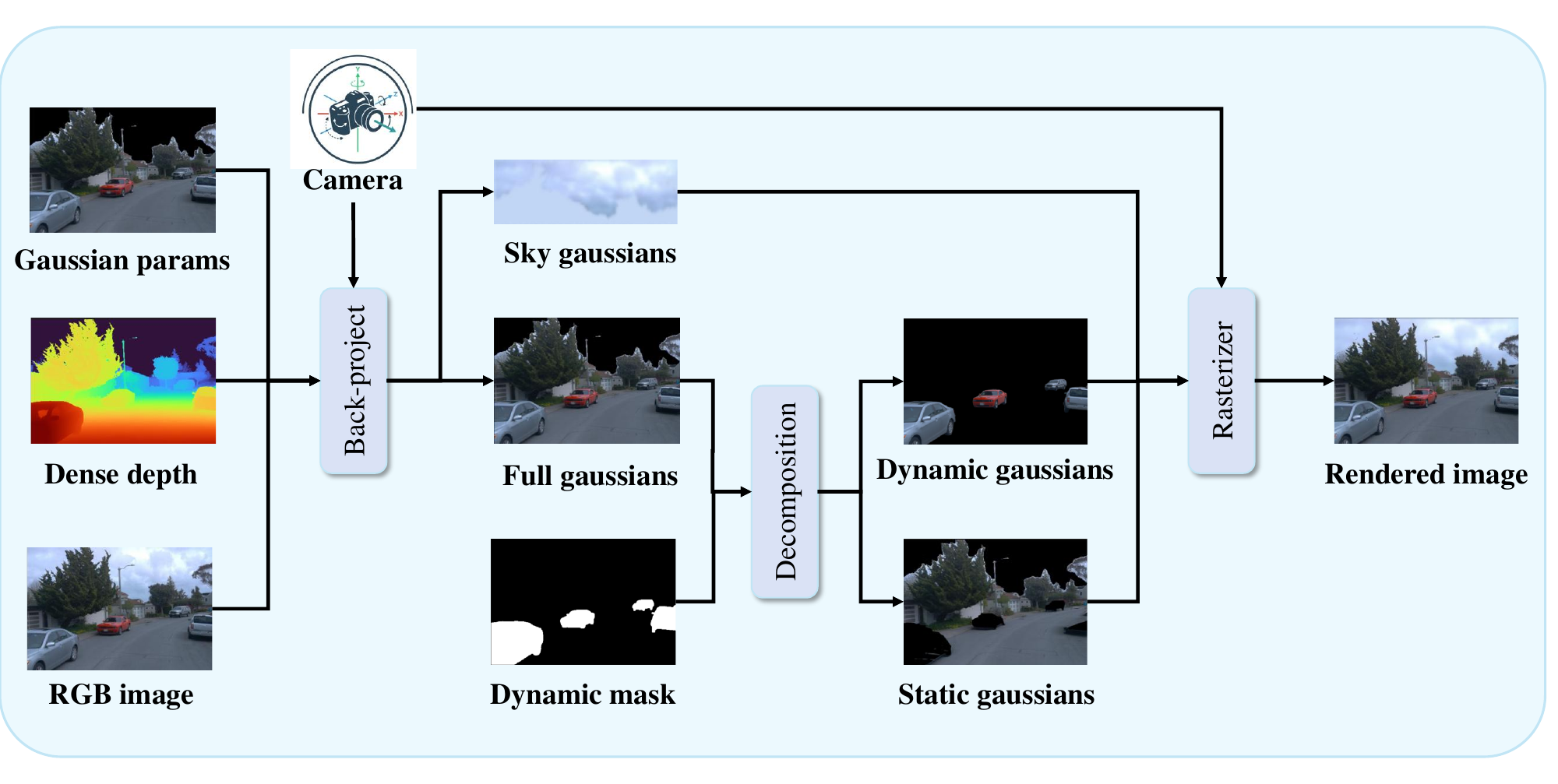}
    \caption{\textbf{Dynamic-static Gaussian rendering.} Dense depth maps are back-projected with camera parameters to obtain Gaussian positions. The predicted dynamic masks allocate traffic-participant regions with high potentially dynamic prior to per-frame Gaussians and the remaining regions to the global static Gaussian set, while sky Gaussians model low-opacity sky regions. All Gaussian components are jointly rendered with a differentiable rasterizer.}
    \label{fig:render}
\end{figure*}

This module takes as input the predicted Gaussian parameters excluding the positions $A^{t}$, the estimated dense depth maps $D^{\mathrm{dense}}_{t}$, the dynamic masks $M^{t}$, the input images $I^{t}$, and the known camera parameters $C^{t}$ from pipeline input. It outputs the rendered dynamic scene $I^{t}_{\mathrm{rendered}}$ from either the input viewpoints or novel viewpoints. An overview of this module is illustrated in Fig.~\ref{fig:render}.

We first back-project the dense depth maps into 3D space using the camera parameters to obtain the Gaussian positions $\mu^{t}$. Compared with directly regressing Gaussian positions, the LiDAR-guided depth estimation provides more accurate geometry and stronger multi-view consistency. The recovered positions are then combined with the predicted Gaussian attributes $A^{t}$ to construct the per-pixel Gaussian representation $G^{t}$ for each frame.

Following DGGT~\cite{chenfeedforward}, we use the predicted dynamic masks to separate per-frame Gaussian primitives from static ones, preventing traffic-participant regions prone to dynamic-static leakage from being absorbed into the global static representation. Specifically, the dynamic Gaussians for each frame, denoted by $G^{t}_{\mathrm{dynamic}}$, and the global static Gaussian set, denoted by $G_{\mathrm{static}}$, are obtained as

\begin{equation}
\begin{aligned}
G^{t}_{\mathrm{dynamic}} &= G^{t}\odot M^{t},\\
G_{\mathrm{static}} &= \bigcup_{t=1}^{T} G^{t}\odot(1-M^{t}),
\end{aligned}
\end{equation}

where $\odot$ denotes masked selection. Consequently, high-prior traffic-participant regions and the static background are explicitly allocated to different Gaussian sets for rendering.

In addition to the ground Gaussians, we introduce a simple yet effective representation for the sky. Specifically, we construct a large hemisphere centered at the camera position of the first viewpoint in the first frame and uniformly sample a set of Gaussian primitives on its surface. The rotation and scale of these sky Gaussians are kept fixed during training and inference.

To initialize the appearance of the sky Gaussians, we first render the ground Gaussians from the input viewpoints and identify image regions with low accumulated opacity. The RGB values of these regions are then projected onto the corresponding sky Gaussians according to the camera parameters $C^{t}$. If a sky Gaussian is never assigned a color during this process, its color is initialized as the average color of all colored sky Gaussians.

Finally, the dynamic Gaussians, the global static Gaussians, and the sky Gaussians are jointly rendered using the Gsplat\cite{ye2024gsplatopensourcelibrarygaussian} rasterizer to synthesize the final image.

\subsection{Instance-Level Semantic Learning}
\label{sec:instance}

To provide a potentially dynamic participant prior for dynamic scene reconstruction, we introduce an instance-level semantic branch that shares the backbone features $F_{\mathrm{attn}}$ with the reconstruction branch. This branch is designed to learn instance-aware semantic representations, which are subsequently used to supervise dynamic mask prediction through the proposed Instance-Dynamic Consistency Constraint.

Following the query-based instance-level semantic representations paradigm, we employ a set of learnable instance query embeddings to progressively extract instance-aware features from different feature levels. Specifically, we adopt the query update strategy of SegVGGT\cite{qu2026segvggt}. Given the feature map $F^{i}$ from the $i$-th attention layer and the corresponding query embeddings $Q^{i}$, the queries are updated through a cross-attention layer followed by a self-attention layer:

\begin{equation}
Q^{i+1}
=
\mathrm{SelfAttn}
\left(
\mathrm{CrossAttn}
(q=Q^{i},k=F^{i},v=F^{i})
\right),
\end{equation}

where $Q^{0}$ denotes the learnable instance query embeddings, and $Q^{N}$ represents the final instance-aware query features obtained after all attention layers.

Meanwhile, we employ a DPT-style\cite{ranftl2021visiontransformersdenseprediction} prediction head to predict a dense feature map for each frame. Let $H_{\mathrm{feat}}$ denote the prediction head and $F^{\mathrm{inst}}$ denote the predicted feature map:

\begin{equation}
F^{\mathrm{inst}}
=
H_{\mathrm{feat}}
(F_{\mathrm{attn}}).
\end{equation}

Finally, the query features $Q^{N}$ are transformed by a multilayer perceptron (MLP) and projected onto the dense feature map $F^{\mathrm{inst}}$ through a dot-product operation, yielding the final instance-level semantic masks.

\subsection{Training Objectives}
\label{sec:train}

We train IGRecon4D in two stages so that the instance-level semantic branch first learns a reliable potentially dynamic participant prior before it is used to supervise reconstruction. In the first stage, we train the instance-level semantic branch with pseudo instance annotations. Following DETR\cite{carion2020endtoendobjectdetectiontransformers}, we employ the Hungarian matching algorithm to establish one-to-one correspondences between the predicted instances and the pseudo instance targets. The first-stage instance loss consists of a classification loss, a binary cross-entropy (BCE) loss, and a Dice loss:

\begin{equation}
\mathcal{L}_{\mathrm{stage1}}
=
\lambda_{\mathrm{cls}}\mathcal{L}_{\mathrm{cls}}
+
\lambda_{\mathrm{inst}}
\left(
\mathcal{L}_{\mathrm{bce}}
+
\mathcal{L}_{\mathrm{dice}}
\right).
\end{equation}

In the second stage, we freeze the instance-level semantic branch trained in the first stage and use its predicted instance masks as a potentially dynamic participant prior for reconstruction. The key idea of the proposed \textbf{Instance-Dynamic Consistency Constraint (IDC)} is to transfer the frozen branch's high-recall localization of traffic-participant regions to the dynamic-mask prediction head. Specifically, let $inst_i^t$ denote the predicted mask of the $i$-th instance at time step $t$, where a total of $K$ instances are predicted. The prior target for the dynamic mask is obtained by taking the union of all predicted instance masks:

\begin{equation}
\tilde{M}^{t}_{\mathrm{inst}}
=
\bigcup_{i=1}^{K} inst_i^t
.
\end{equation}

The prior-guided dynamic mask loss is then defined as

\begin{equation}
\mathcal{L}_{\mathrm{dynamic}}
=
\mathrm{BCE}
\left(
M^{t},
\tilde{M}^{t}_{\mathrm{inst}}
\right).
\end{equation}

The second stage optimizes the Gaussian prediction head and the dynamic-mask prediction head for reconstruction, while keeping the first-stage modules fixed. To supervise Gaussian parameter prediction, we introduce a sky loss that encourages Gaussian primitives corresponding to the sky to remain transparent. The sky loss is formulated as

\begin{equation}
\mathcal{L}_{\mathrm{sky}}
=
\mathrm{BCE}
\left(
\{M_{\mathrm{sky}}^{t}\},
\{\hat{M}_{\mathrm{sky}}^{t}\}
\right),
\end{equation}

where $\{M_{\mathrm{sky}}^{t}\}$ denotes the predicted sky mask obtained from low-opacity regions of the rendered image, and $\{\hat{M}_{\mathrm{sky}}^{t}\}$ is the ground-truth sky mask.

To regularize the temporal persistence of Gaussian primitives, we impose an $\ell_{1}$ regularization on the inverse lifetime parameter:

\begin{equation}
\mathcal{L}_{\mathrm{reg}}
=
\left\|
\frac{1}{\sigma}
\right\|_{1}.
\end{equation}

Furthermore, we supervise the rendered images using both an image reconstruction loss and a perceptual loss. Specifically, an $\ell_{2}$ loss and an LPIPS\cite{zhang2018perceptual} loss are computed between the rendered image $\hat{I}^{t}$ and the ground-truth image $I^{t}$.

The second-stage reconstruction objective is defined as

\begin{equation}
\begin{aligned}
\mathcal{L}_{\mathrm{stage2}}
=&\;
\mathcal{L}_{\ell_2}(I^{t},\hat{I}^{t})
+\lambda_{\mathrm{lpips}}
\mathcal{L}_{\mathrm{LPIPS}}(I^{t},\hat{I}^{t})\\
&+\lambda_{\mathrm{dynamic}}
\mathcal{L}_{\mathrm{dynamic}}
+\lambda_{\mathrm{reg}}
\mathcal{L}_{\mathrm{reg}}
+\lambda_{\mathrm{sky}}
\mathcal{L}_{\mathrm{sky}}.
\end{aligned}
\end{equation}

\section{Experiments}

\subsection{Datasets and Evaluation Protocol}

We conduct comprehensive experiments on the Waymo Open Dataset\cite{jund2021scalablesceneflowpoint}. The dataset covers a wide variety of driving scenarios, including urban streets, highways, and complex traffic environments, and contains abundant traffic participants (e.g., pedestrians, vehicles, and cyclists) under diverse illumination and weather conditions. These characteristics make it well suited for evaluating the generalization ability and robustness of dynamic scene reconstruction methods in challenging real-world scenarios.

During training, we use the entire training split, which consists of 798 scenes. Each scene contains approximately 190--200 consecutive frames, providing sufficient temporal information for learning dynamic scene representations. During evaluation, we report results on the validation split containing 202 scenes. Unless otherwise specified, all reconstruction methods are evaluated under the same validation split, input camera configuration, and metric computation protocol.

Before training and inference, we perform several preprocessing steps. To obtain instance-level semantic annotations, we employ SAM3\cite{carion2025sam3segmentconcepts} to automatically generate instance masks for each scene. Specifically, language prompts corresponding to traffic participant categories, including \emph{pedestrian}, \emph{vehicle}, and \emph{cyclist}, are used to extract the corresponding instances. These masks are used as a potentially dynamic participant prior rather than as ground-truth motion-state labels. In addition, we generate sky masks for every frame using LISA\cite{lai2023lisa}. Benefiting from its language-guided semantic segmentation capability, LISA accurately identifies large semantic regions with weak texture, such as the sky, and produces reliable sky masks for supervision during training.

\subsection{Implementation Details}

Our training procedure consists of two stages. In the first stage, we train the instance-level semantic branch to learn instance-aware semantic representations. In the second stage, we leverage the learned semantic representations to improve dynamic scene reconstruction through the proposed Instance-Dynamic Consistency Constraint. Both stages are trained on eight NVIDIA B200 GPUs. Following DGGT\cite{chenfeedforward}, we adopt the same data sampling strategy by uniformly sampling four frames from every consecutive 20-frame sequence captured by the three front-facing cameras. In both stages, the AdamW\cite{loshchilov2019decoupledweightdecayregularization} optimizer is employed together with a linear warm-up schedule followed by cosine learning rate decay.

In the first stage, the batch size is set to 8 per GPU, and the model is trained for 10,000 iterations. We initialize the backbone with the pretrained VGGT\cite{wang2025vggt} weights and perform LoRA\cite{hu2021loralowrankadaptationlarge} fine-tuning (rank = 32) on the alternating attention layers. Meanwhile, the cross-attention layers, self-attention layers, MLPs, and the instance prediction head are fully fine-tuned. We use 100 learnable instance query embeddings in the instance-level semantic branch. The learning rates are set to $6 \times 10^{-5}$ for the LoRA parameters and $2 \times 10^{-4}$ for the fully trainable parameters. We set $\lambda_{\mathrm{cls}}=0.5$ and $\lambda_{\mathrm{inst}}=1.0$ in this stage.

In the second stage, the batch size is reduced to 1 per GPU, and the model is trained for 50,000 iterations. The depth predictor is initialized with the pretrained InfiniDepth\cite{yu2026infinidepth} weights. We freeze the LoRA-adapted alternating attention layers as well as the modules trained in the first stage, and optimize only the Gaussian prediction head and the dynamic prediction head. The learning rate in this stage is set to $4 \times 10^{-5}$. We set $\lambda_{\mathrm{lpips}}=0.05$, $\lambda_{\mathrm{dynamic}}=0.05$, $\lambda_{\mathrm{reg}}=0.01$ and $\lambda_{\mathrm{sky}}=1.0$ in this stage.

During inference, we use 20 consecutive frames captured by the three front-facing cameras as input and evaluate the performance of our method. We conduct our evaluation on a single NVIDIA RTX Pro 6000 GPU.

\subsection{Comparison with State-of-the-Art Methods}

We compare the proposed method with several representative feed-forward reconstruction approaches to comprehensively evaluate its performance on dynamic scene reconstruction. Specifically, the compared feed-forward methods include AnySplat\cite{jiang2025anysplat}, STORM\cite{yang2025storm}, DGGT\cite{chenfeedforward}, NoPo4D\cite{balice2026poseproblem4dfeedforward}, and DrivingForward\cite{tian2025drivingforward}. For a fair comparison, all methods are evaluated using the same number of input frames, namely 20 consecutive frames captured by the three front-facing cameras. Comparisons with additional methods are provided in Appendix~\ref{app:more_comparison}.

\begin{table*}[t]
\centering
\caption{\textbf{Quantitative comparison on the Waymo Open Dataset.} Metrics are computed on the validation split under the same input-view and input-frame setting for all methods. Traffic-participant-region (TPR) metrics are computed on SAM3-generated traffic-participant masks, while full-image metrics are computed over all pixels.}
\label{tab:comparison}
\begingroup
\small
\setlength{\tabcolsep}{3pt}
\begin{tabular}{lcccccccc}
\toprule
Method & TPR PSNR $\uparrow$ & TPR SSIM $\uparrow$ & PSNR $\uparrow$ & SSIM $\uparrow$ & LPIPS $\downarrow$ & D-RMSE $\downarrow$ & Time (s) $\downarrow$ & Memory (GB) $\downarrow$ \\
\midrule
AnySplat\cite{jiang2025anysplat}  & 21.70 & 0.680 & 25.43 & 0.751 & 0.217 & 9.756 & 2.56 & 20.96 \\
STORM\cite{yang2025storm}     & 19.25 & 0.504 & 25.37 & 0.776 & 0.315 & 18.849 & 1.74 & 85.35 \\
DGGT\cite{chenfeedforward}    & 24.82 & 0.751  & 26.93 & 0.868 & 0.116 & 7.470 & 10.88 & 47.36 \\
NoPo4D\cite{balice2026poseproblem4dfeedforward}    & 21.70 & 0.640  & 25.27 & 0.776 & 0.200 & 8.872 & 3.61 & 24.00 \\
DrivingForward\cite{tian2025drivingforward}   & 19.55 & 0.751 & 17.61 & 0.802 & 0.222 & 12.007 & 2.73 & 2.78 \\
Ours      & \textbf{26.76} & \textbf{0.858} & \textbf{29.51} & \textbf{0.869} & \textbf{0.102} & \textbf{4.610} & 16.20 & 69.27 \\
\bottomrule
\end{tabular}
\endgroup
\end{table*}

For quantitative evaluation, we report image quality, geometric consistency, and computational cost. Image quality is measured using PSNR, SSIM, and LPIPS\cite{zhang2018perceptual}. As shown in Table~\ref{tab:comparison}, our method improves PSNR over the strongest baseline DGGT\cite{chenfeedforward} by 2.58 dB and reduces LPIPS from 0.116 to 0.102, while also achieving a slightly higher SSIM. To assess geometric consistency with LiDAR observations, we additionally report the average per-scene depth root mean square error (D-RMSE). Following a protocol similar to DGGT\cite{chenfeedforward}, we linearly align the depths rendered from the reconstructed Gaussians to sparse depths obtained by projecting LiDAR points, compute RMSE at pixels with valid sparse depth, and average the resulting per-scene errors. Our method achieves the lowest D-RMSE of 4.610, compared with 7.470 for DGGT. We also report the runtime and GPU memory usage of each method. Although our method does not have the lowest runtime or memory usage, it completes feed-forward dynamic Gaussian reconstruction in 16.20 s.

\begin{figure*}[t]
    \centering
    \includegraphics[width=\textwidth]{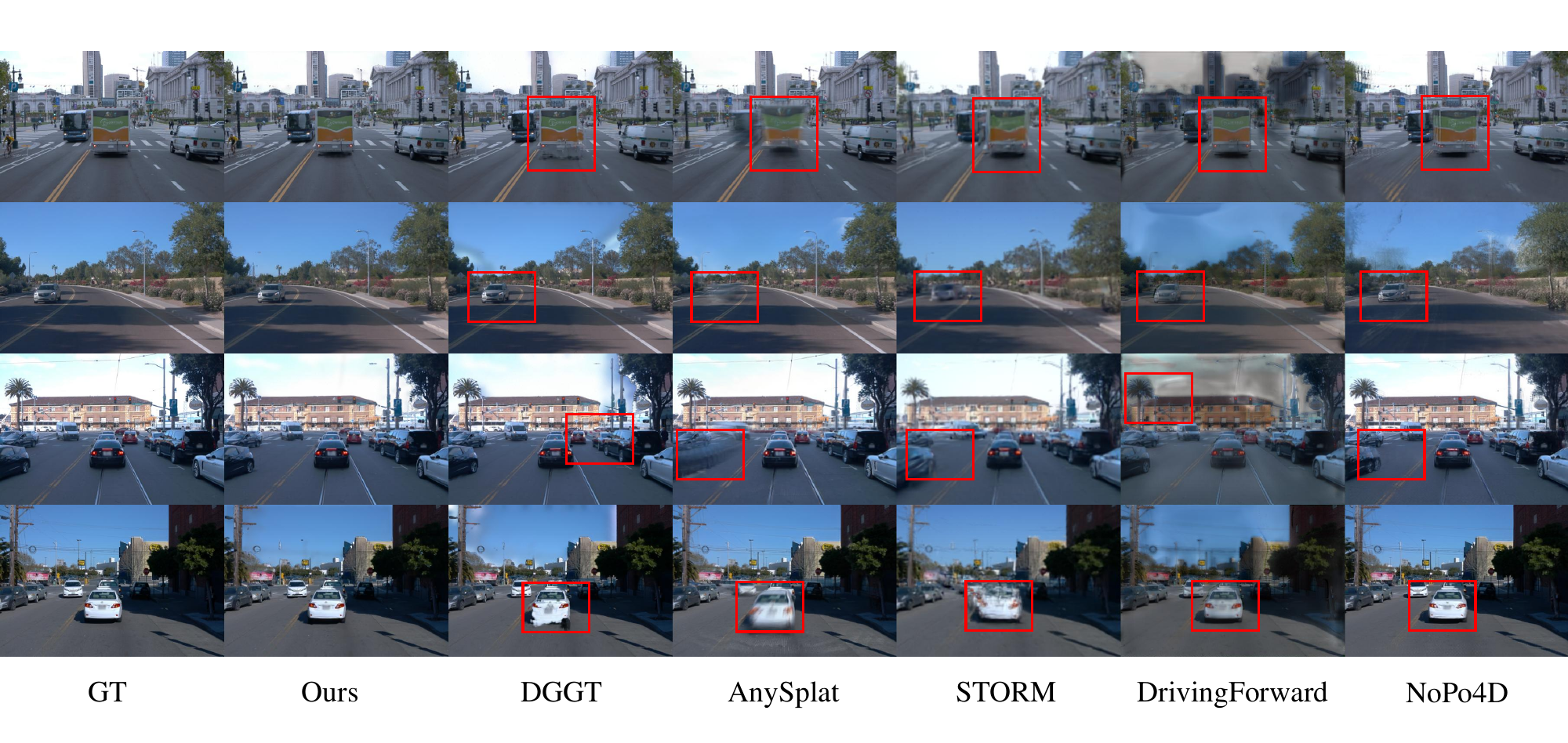}
    \caption{\textbf{Qualitative comparison on the Waymo Open Dataset.} IGRecon4D better preserves traffic participants and static background details than feed-forward baselines, with fewer participant-region omissions and sharper object boundaries.}
    \label{fig:comparison}
\end{figure*}

We further conduct qualitative comparisons with existing feed-forward reconstruction methods. As illustrated in Fig.~\ref{fig:comparison}, the compared methods exhibit noticeable differences in traffic-participant-region reconstruction and texture preservation. Specifically, DGGT\cite{chenfeedforward} provides insufficient high-recall localization of traffic-participant regions that are prone to dynamic-static leakage, causing some pedestrians or vehicles to be absorbed into the static representation and degrading reconstruction quality. STORM\cite{yang2025storm}, DrivingForward\cite{tian2025drivingforward} and NoPo4D\cite{balice2026poseproblem4dfeedforward} exhibit noticeable information loss during texture modeling, resulting in blurred surface details. Although AnySplat\cite{jiang2025anysplat} achieves competitive performance on static scenes, its lack of explicit per-frame participant-region modeling makes it difficult to faithfully reconstruct traffic participants in dynamic environments.

In contrast, our method derives more reliable Gaussian positions through LiDAR-guided depth estimation while better localizing traffic-participant regions that should be modeled by per-frame Gaussians. Consequently, it improves the reconstruction of both static backgrounds and traffic participants, such as pedestrians and vehicles, within a unified framework. This leads to clearer object boundaries and better geometric alignment, as further demonstrated by the qualitative results in Fig.~\ref{fig:comparison}.

\subsection{Novel View Synthesis}

To further evaluate the multi-view consistency of the proposed method, we conduct a novel-view synthesis experiment. Specifically, during inference, we reconstruct the dynamic scene from videos captured by three synchronized cameras and obtain the corresponding dynamic Gaussian representation. Instead of rendering from the input viewpoints, we synthesize images from interpolated viewpoints. We denote the front-facing camera as Camera 0, the front-left camera as Camera 1, and the front-right camera as Camera 2. Novel views are rendered from the interpolated viewpoints between Camera 0 and Camera 1, as well as between Camera 0 and Camera 2, to evaluate the reconstruction quality under unseen viewpoints. As a baseline, DGGT is evaluated under the same setting by giving the same camera parameters as input and rendering from the identical interpolated viewpoints. 

\begin{figure*}[t]
    \centering
    \includegraphics[width=\textwidth]{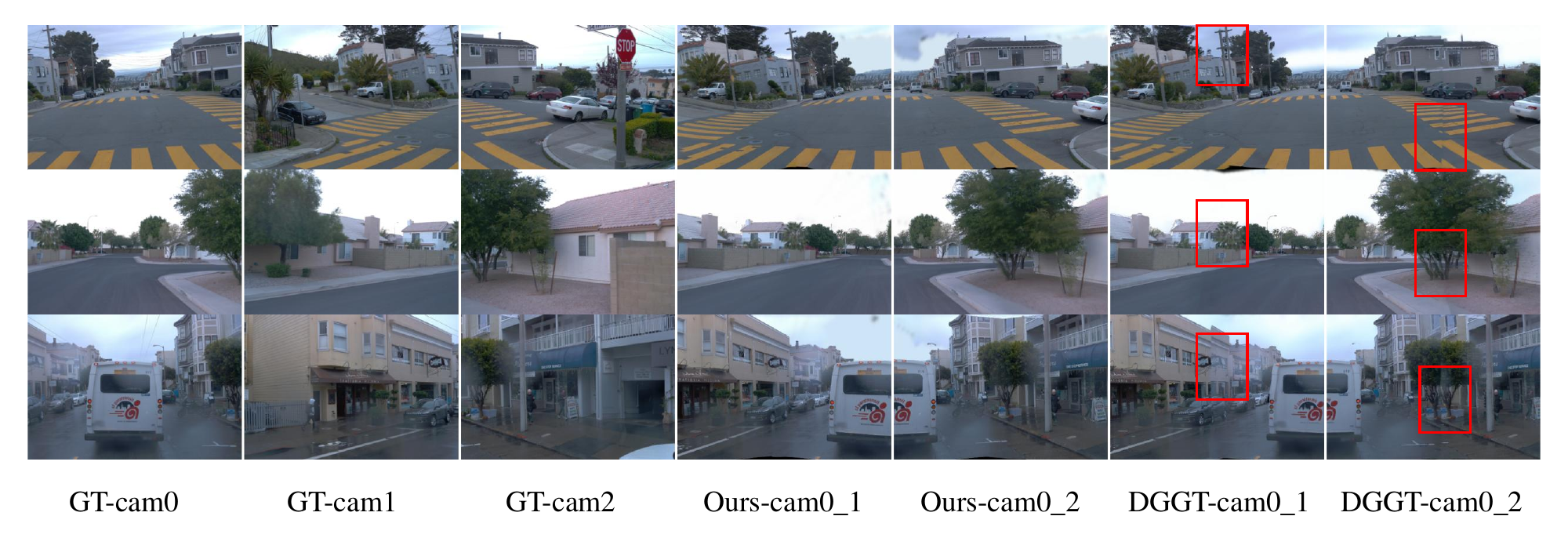}
    \caption{
        \textbf{Novel-view synthesis from interpolated camera poses.}
        We render from interpolated viewpoints between Camera~0 and Camera~1,
        and between Camera~0 and Camera~2, and compare IGRecon4D with
        DGGT~\cite{chenfeedforward} under identical camera poses.
        \texttt{cam0\_1} and \texttt{cam0\_2} denote the interpolated viewpoints
        between Camera~0 and Camera~1, and between Camera~0 and Camera~2,
        respectively. IGRecon4D produces fewer geometric shifts and stronger
        multi-view consistency.
    }
    \label{fig:novel_view}
\end{figure*}

Qualitative comparisons are presented in Fig.~\ref{fig:novel_view}. As can be observed, DGGT\cite{chenfeedforward} suffers from inaccurate depth estimation, causing noticeable geometric misalignment and spatial shifts of objects that should remain geometrically consistent under novel viewpoints. In contrast, our LiDAR-guided depth estimation derives more reliable Gaussian positions for LiDAR-anchored Gaussian positioning, resulting in improved geometric alignment and stronger structural consistency under novel-view rendering.

\begin{table}[t]
\centering
\caption{\textbf{Novel-view synthesis measured by FID.} FID is used as a proxy metric between rendered interpolated views and the real-image distribution under the same evaluation protocol; lower scores indicate smaller distributional discrepancy.}
\label{tab:novel_view}
\begin{tabular}{lcc}
\toprule
Method & Left View FID $\downarrow$ & Right View FID $\downarrow$ \\
\midrule
Ours & \textbf{117.73} & \textbf{132.41} \\
DGGT\cite{chenfeedforward} & 129.47 & 151.10 \\
\bottomrule
\end{tabular}
\end{table}

Quantitative results are reported in Table~\ref{tab:novel_view}. Because exact ground-truth images at the interpolated camera poses are unavailable, we use the Fr\'echet Inception Distance (FID) as a proxy metric and compute the distributional distance between rendered interpolated-view images and real images from the input-view distribution. As shown in Table~\ref{tab:novel_view}, our method achieves lower FID scores than DGGT\cite{chenfeedforward} under this protocol, suggesting better distributional consistency for interpolated-view rendering.

Overall, these results suggest that more accurate depth estimation improves the spatial localization of Gaussian primitives, leading to stronger multi-view geometric consistency and higher-quality novel-view synthesis.

\subsection{Ablation Study}

The improvements in reconstruction quality mainly stem from two key components: the proposed \textbf{Instance-Dynamic Consistency Constraint (IDC)} and the LiDAR-guided depth estimation module. We therefore conduct targeted ablation studies to assess the contribution of these components through both quantitative metrics and qualitative visual comparisons. Additional ablation results are provided in Appendix~\ref{app:more_ablation}.

\begin{table*}[t]
\centering
\caption{\textbf{Ablation study on the Waymo Open Dataset.} Traffic-participant-region (TPR) metrics are computed on SAM3-generated traffic-participant masks, while full-image metrics are computed over all pixels.}
\label{tab:ablation}
\begin{tabular}{lcccccc}
\toprule
Method & TPR PSNR $\uparrow$ & TPR SSIM $\uparrow$ & PSNR $\uparrow$ & SSIM $\uparrow$ & LPIPS $\downarrow$ & D-RMSE $\downarrow$ \\
\midrule
Ours & \textbf{26.76} & \textbf{0.858} & \textbf{29.51} & \textbf{0.869} & \textbf{0.102} & \textbf{4.610} \\
w/o IDC & 23.81 & 0.774 & 28.92 & 0.862 & 0.108 & 4.624 \\
w/o LiDAR-guided depth & 23.73 & 0.758 & 25.20 & 0.759 & 0.213 & 11.584 \\
w/o IDC and w/o LiDAR-guided depth & 21.02 & 0.626 & 24.77 & 0.749 & 0.222 & 11.606 \\
\bottomrule
\end{tabular}
\end{table*}

\begin{figure*}[t]
    \centering
    \includegraphics[width=\linewidth]{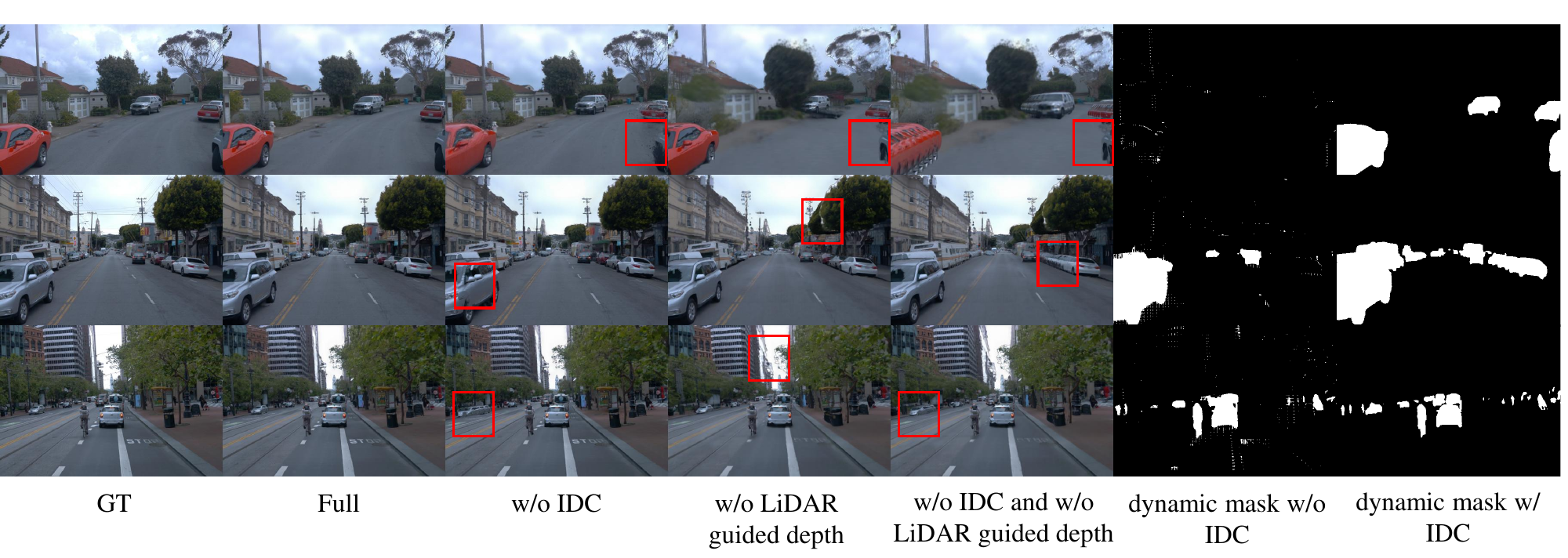}
    \caption{\textbf{Qualitative ablation study.} Replacing IDC-supervised dynamic-mask learning with a DGGT-initialized dynamic head reduces high-recall localization of traffic-participant regions, while replacing LiDAR-guided depth estimation with DPT-style depth prediction introduces geometric misalignment and degrades rendering quality.}
    \label{fig:ablation}
\end{figure*}

We first evaluate the role of IDC in dynamic-mask learning. For the \emph{w/o IDC} variant, we remove the instance-level semantic branch and the IDC loss. Instead, the dynamic-mask prediction head is initialized from the pretrained dynamic head of DGGT. This setting provides a strong DGGT-style dynamic-mask baseline under the same input, sampling, rendering, and reconstruction-loss protocol. Therefore, this variant evaluates whether the proposed potentially dynamic participant prior improves traffic-participant-region reconstruction beyond a pretrained dynamic-mask predictor.

As shown in Table~\ref{tab:ablation} and Fig.~\ref{fig:ablation}, replacing IDC-supervised dynamic-mask learning with the DGGT-initialized dynamic head reduces TPR (traffic-participant-region) PSNR from 26.76 to 23.81 and TPR SSIM from 0.858 to 0.774. This suggests that a pretrained dynamic head alone is insufficient for high-recall localization of traffic-participant regions that are prone to dynamic-static leakage in driving scenes. In contrast, IDC transfers the potentially dynamic participant prior to dynamic-mask prediction, leading to improved traffic-participant-region reconstruction.

We further evaluate the contribution of the proposed LiDAR-guided depth estimation module. The \emph{w/o LiDAR-guided depth} variant replaces LiDAR-guided depth estimation with the DPT-style\cite{ranftl2021visiontransformersdenseprediction} dense-depth prediction strategy adopted in DGGT\cite{chenfeedforward}, while keeping all other modules, input camera parameters and training settings unchanged. This variant isolates whether LiDAR-anchored Gaussian positioning improves reconstruction quality.

The quantitative and qualitative comparisons demonstrate that replacing LiDAR-guided depth estimation with a DPT-style predictor substantially degrades reconstruction performance, reducing full-image PSNR from 29.51 to 25.20 and increasing LPIPS from 0.102 to 0.213. The degradation is especially noticeable around object boundaries and regions with complex occlusions, where inaccurate depth estimation accumulates geometric errors and deteriorates the final rendering quality. These results indicate that LiDAR-guided depth estimation provides more accurate and geometrically consistent Gaussian positions.

Overall, the ablation studies validate the two main design claims of IGRecon4D. IDC primarily improves high-recall localization of traffic-participant regions by transferring the potentially dynamic participant prior to dynamic-mask learning, whereas LiDAR-guided depth estimation mainly improves LiDAR-anchored Gaussian positioning and full-scene reconstruction quality. Both the quantitative results in Table~\ref{tab:ablation} and the qualitative comparisons in Fig.~\ref{fig:ablation} show that these components address complementary failure modes.

\subsection{Zero-shot Evaluation on nuScenes}

To further evaluate the generalization capability of the proposed method, we conduct zero-shot experiments on the nuScenes\cite{caesar2020nuscenesmultimodaldatasetautonomous} dataset. We compare our method with several representative feed-forward reconstruction approaches, including AnySplat\cite{jiang2025anysplat}, STORM\cite{yang2025storm}, DGGT\cite{chenfeedforward}, NoPo4D\cite{balice2026poseproblem4dfeedforward} and DrivingForward\cite{tian2025drivingforward}.

\begin{table*}[t]
\centering
\caption{\textbf{Zero-shot reconstruction comparison on nuScenes.} Models are evaluated without fine-tuning on nuScenes; traffic-participant-region (TPR) metrics are computed on traffic-participant masks.}
\label{tab:zeroshot}
\begingroup
\small
\begin{tabular}{lccccc}
\toprule
Method & TPR PSNR $\uparrow$ & TPR SSIM $\uparrow$  & PSNR $\uparrow$ & SSIM $\uparrow$ & LPIPS $\downarrow$\\
\midrule
AnySplat\cite{jiang2025anysplat}  & 20.33 & 0.545 & \textbf{27.89} & \textbf{0.857} & \textbf{0.120} \\
STORM\cite{yang2025storm}     & 15.57 & 0.255 & 16.90 & 0.472 & 0.533 \\
DGGT\cite{chenfeedforward}      & 23.43 & 0.722 & 20.73 & 0.807 & 0.173 \\
NoPo4D\cite{balice2026poseproblem4dfeedforward}    & 20.22 & 0.598 & 23.10 & 0.655 & 0.282 \\
DrivingForward\cite{tian2025drivingforward}   & 18.88 & 0.764 & 18.89 & 0.775 & 0.191 \\
Ours      & \textbf{25.05} & \textbf{0.809} & 25.93 & 0.752 & 0.172 \\
\bottomrule
\end{tabular}
\endgroup
\end{table*}

\begin{figure*}[t]
    \centering
    \includegraphics[width=\textwidth]{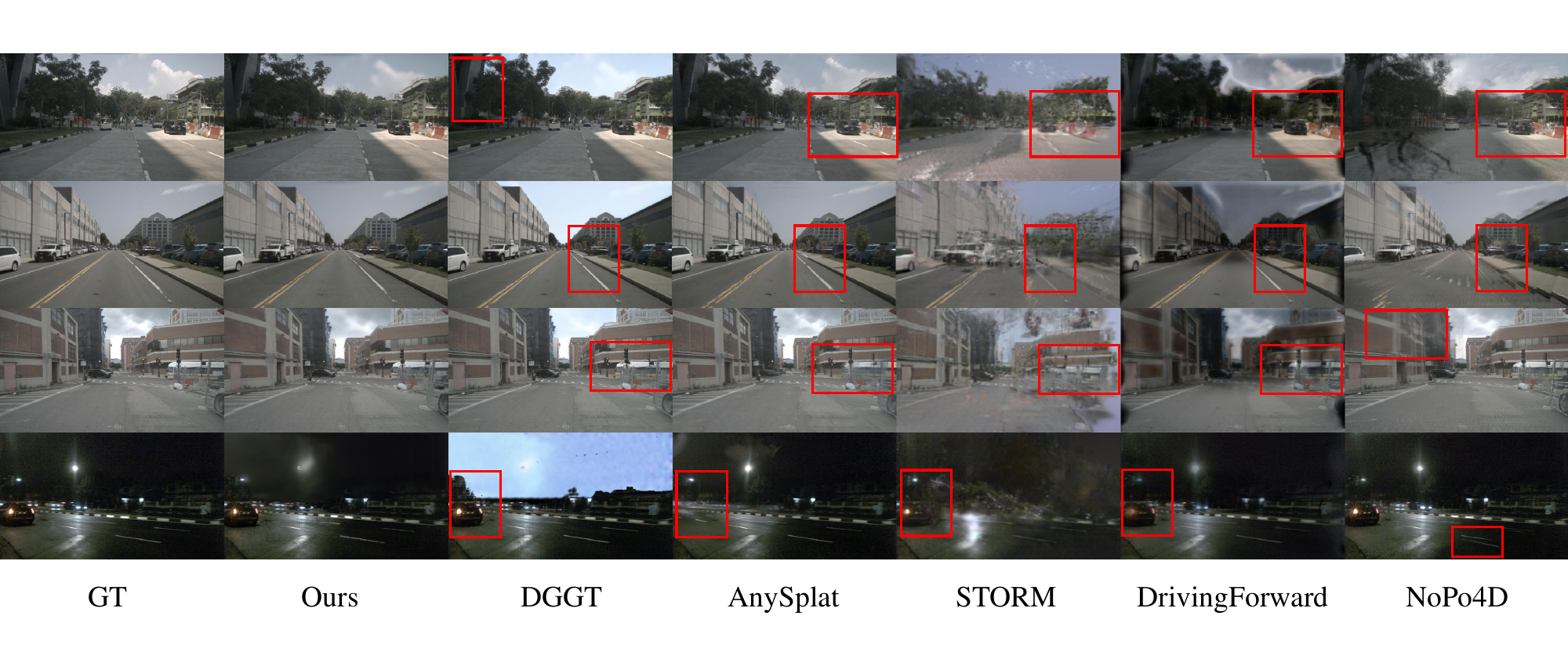}
    \caption{\textbf{Qualitative zero-shot reconstruction on nuScenes.} Without fine-tuning on nuScenes, IGRecon4D preserves traffic participants more reliably than feed-forward baselines.}
    \label{fig:zeroshot}
\end{figure*}

The quantitative results are reported in Table~\ref{tab:zeroshot}. Our method achieves the best TPR PSNR and TPR SSIM, suggesting stronger cross-dataset traffic-participant-region reconstruction. AnySplat obtains better full-image PSNR, SSIM, and LPIPS, which suggests an advantage on static or background-dominated regions; however, its TPR metrics are substantially lower because it lacks an explicit mechanism for modeling per-frame participant regions. STORM, NoPo4D and DrivingForward suffer from noticeable texture degradation, while DGGT remains less accurate on traffic-participant regions. Our zero-shot visualization results on nuScenes are shown in Fig.~\ref{fig:zeroshot}.

Overall, IGRecon4D achieves the strongest zero-shot traffic-participant-region reconstruction quality on the unseen nuScenes dataset. These results suggest that the proposed method generalizes most clearly to traffic participants, which are the main regions prone to dynamic-static leakage in autonomous-driving reconstruction.

\subsection{Agreement with the Pseudo-label Teacher}

We further evaluate whether the instance-level semantic branch provides a reliable potentially dynamic participant prior for IDC. This evaluation is not intended to establish a new state of the art in instance segmentation; instead, it verifies that the instance-level semantic branch can produce semantic masks of sufficient quality to supervise dynamic-mask learning. We compare the instance-level semantic branch with representative instance segmentation methods, including Mask2Former\cite{cheng2021mask2former}, DVIS++\cite{zhang2023dvisimproveddecoupledframework}, and VidEoMT\cite{norouzi2026videomt}.

Since the Waymo Open Dataset\cite{jund2021scalablesceneflowpoint} does not provide manually annotated instance segmentation labels, we adopt the instance masks generated by SAM3 as pseudo instance annotations. We report mAP, mAP$_{50}$, mAP$_{25}$, PQ, and mIoU with respect to these pseudo annotations. Following the standard PQ evaluation protocol, pixels that do not belong to any instance are treated as the background class. The quantitative results are summarized in Table~\ref{tab:instance}.

\begin{table}[t]
\centering
\caption{\textbf{Agreement with the Pseudo-label Teacher.} All methods are evaluated against SAM3-generated pseudo instance annotations on Waymo; the branch is used to provide IDC supervision rather than to serve as a standalone segmentation model.}
\label{tab:instance}
\begingroup
\setlength{\tabcolsep}{2pt}
\footnotesize
\begin{tabular}{lccccc}
\toprule
Method & mAP $\uparrow$ & mAP$_{50}$ $\uparrow$ & mAP$_{25}$ $\uparrow$ & PQ $\uparrow$ & mIoU $\uparrow$ \\
\midrule
Ours        & 0.540 & 0.808 & 0.898 & 0.574 & 0.918 \\
Mask2Former\cite{cheng2021mask2former} & 0.660 & 0.843 & 0.872 & 0.174 & 0.848 \\
DVIS++\cite{zhang2023dvisimproveddecoupledframework}      & 0.686 & 0.895 & 0.913 & 0.612 & 0.926 \\
VidEoMT\cite{norouzi2026videomt}     & 0.630 & 0.886 & 0.904 & 0.567 & 0.915 \\
\bottomrule
\end{tabular}
\endgroup
\end{table}

As shown in Table~\ref{tab:instance}, specialized segmentation models can achieve higher mAP or PQ under this pseudo-label protocol. Nevertheless, the proposed instance-level semantic branch obtains competitive mIoU and produces coherent instance masks for traffic participants. More importantly, the ablation results in Table~\ref{tab:ablation} show that using these masks through IDC substantially improves traffic-participant-region reconstruction, confirming their effectiveness as a prior rather than as a standalone segmentation output.

\begin{figure*}[t]
    \centering
    \includegraphics[width=\textwidth]{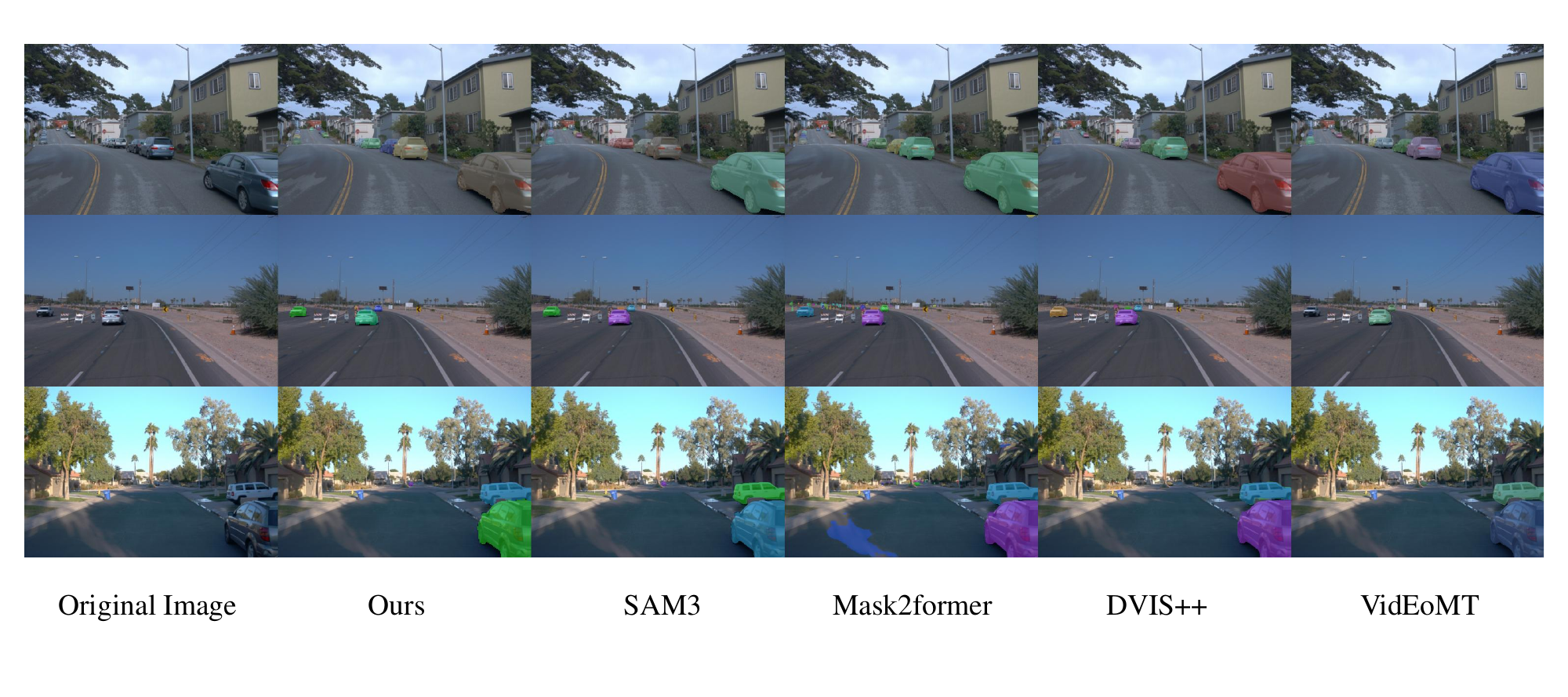}
    \caption{\textbf{Instance-level semantic predictions.} The instance-level semantic branch is evaluated as a source of a potentially dynamic participant prior for IDC rather than as a standalone segmentation model; the predicted masks provide usable instance-level supervision for dynamic-mask learning.}
    \label{fig:instance}
\end{figure*}

Qualitative comparisons of the predicted instance masks are presented in Fig.~\ref{fig:instance}, further showing that the learned semantic representations are sufficiently accurate to facilitate dynamic scene reconstruction.

\section{Conclusion}

In this paper, we presented IGRecon4D, a feed-forward framework for dynamic scene reconstruction in autonomous driving scenarios. The key idea is to leverage instance-level semantic guidance and LiDAR-guided depth estimation in a feed-forward Gaussian reconstruction pipeline: the instance-level semantic branch provides a potentially dynamic participant prior for mask supervision, while LiDAR-guided depth estimation derives reliable Gaussian positions for LiDAR-anchored Gaussian positioning.

Experiments on the Waymo Open Dataset show that IGRecon4D improves full-image reconstruction quality over representative feed-forward baselines and yields clear gains on traffic-participant regions in the ablation study. Novel-view synthesis further supports the benefit of LiDAR-anchored Gaussian positioning for multi-view consistency. Under zero-shot evaluation on nuScenes, IGRecon4D achieves the best traffic-participant-region reconstruction metrics, although full-image metrics remain influenced by static and background-dominated regions.

The current framework is evaluated mainly with three front-facing cameras, sparse LiDAR input, and short video clips. Extending the method to longer temporal windows, broader camera configurations, and larger-scale scenes is an important direction for future work. In addition, the instance-level prior is semantic rather than motion-state supervision, and therefore cannot distinguish static and moving objects of the same category, such as parked and moving vehicles. Reducing training and inference cost would further improve its practicality for large-scale autonomous-driving reconstruction.

\appendix

\section*{Appendix}

\begingroup
\raggedbottom
\titleformat{\subsection}{\bfseries\fontsize{10.5pt}{13.5pt}\selectfont}
{\thesubsection}{.8em}{#1}

\section{More implementation details}
\setcounter{equation}{0}
\renewcommand{\theequation}{\thesection\arabic{equation}}
\renewcommand{\theHequation}{appendix.\thesection.\arabic{equation}}

\subsection{More details of LiDAR projection}
\label{app:lidar_projection}

Let $L_t=\{\mathbf{p}_{\ell,i}^{t}\}_{i=1}^{N_t}$ denote the LiDAR points acquired at time $t$. For camera $c$, let $\mathbf{K}_c$ be its intrinsic matrix and $(\mathbf{R}_{c\leftarrow\ell}^{t},\mathbf{t}_{c\leftarrow\ell}^{t})$ be the rigid transform from the LiDAR coordinate frame to the camera coordinate frame. These calibration parameters are included in the known camera parameters $C^t$. Each LiDAR point is first transformed into camera coordinates:
\begin{equation}
\mathbf{p}_{c,i}^{t}
=\mathbf{R}_{c\leftarrow\ell}^{t}\mathbf{p}_{\ell,i}^{t}
+\mathbf{t}_{c\leftarrow\ell}^{t}.
\label{eq:appendix_lidar_transform}
\end{equation}
The point is then projected onto the image plane by
\begin{equation}
\begin{aligned}
(\tilde{u}_{t,i},\tilde{v}_{t,i},\tilde{w}_{t,i})^{\top}
&=\mathbf{K}_c\mathbf{p}_{c,i}^{t},\\
(u_{t,i},v_{t,i})
&=\left(\frac{\tilde{u}_{t,i}}{\tilde{w}_{t,i}},
\frac{\tilde{v}_{t,i}}{\tilde{w}_{t,i}}\right).
\end{aligned}
\label{eq:appendix_lidar_projection}
\end{equation}
Only points with positive camera-frame depth $[\mathbf{p}_{c,i}^{t}]_3$ and projected coordinates inside the image are retained. Let $\mathcal{Z}_t(u,v)$ be the set of their camera-frame depths assigned to pixel $(u,v)$ after pixel discretization. The sparse depth map for camera $c$ (with the camera index suppressed) is
\begin{equation}
D_{t}^{\mathrm{sparse}}(u,v)=
\begin{cases}
\min\mathcal{Z}_t(u,v), & \mathcal{Z}_t(u,v)\ne\varnothing,\\
0, & \text{otherwise}.
\end{cases}
\label{eq:appendix_sparse_depth}
\end{equation}
Keeping the nearest positive depth resolves multiple LiDAR points projecting to the same pixel, while zero marks pixels without a valid LiDAR measurement. The resulting $D_{t}^{\mathrm{sparse}}$ is used as input to the dense depth estimator described in Sec.~\ref{sec:feedforward}.

\section{More experimental results}
\setcounter{equation}{0}

\subsection{Comparison with More Methods}
\label{app:more_comparison}

We extend the comparison in Sec.~4.3 to EVolSplat4D~\cite{miao2026evolsplat4defficientvolumebasedgaussian}, IGGT~\cite{li2025iggt}, and EPS3D~\cite{zhu2026eps3dendtoendfeedforward3d} on the Waymo Open Dataset. EVolSplat4D takes both RGB and LiDAR as input, while IGGT and EPS3D jointly predict scene geometry and instance-level semantic information. Under the same RGB and LiDAR input conditions, our method outperforms EVolSplat4D on the reconstruction and geometric-consistency metrics in Table~\ref{tab:more_comparison}. IGGT and EPS3D also obtain lower reconstruction quality in this driving-scene evaluation, suggesting that their predicted instance-level information does not sufficiently enhance geometric reconstruction in this setting. Fig.~\ref{fig:more_comparison} provides qualitative comparisons.

\begin{table*}[t]
\centering
\caption{\textbf{Additional quantitative comparison on the Waymo Open Dataset.} Metrics are computed on the validation split under the same input-view and input-frame setting for all methods. Traffic-participant-region (TPR) metrics are computed on SAM3-generated traffic-participant masks, while full-image metrics are computed over all pixels.}
\label{tab:more_comparison}
\begingroup
\small
\setlength{\tabcolsep}{3pt}
\begin{tabular}{lcccccccc}
\toprule
Method & TPR PSNR $\uparrow$ & TPR SSIM $\uparrow$ & PSNR $\uparrow$ & SSIM $\uparrow$ & LPIPS $\downarrow$ & D-RMSE $\downarrow$ & Time (s) $\downarrow$ & Memory (GB) $\downarrow$ \\
\midrule
Ours & \textbf{26.76} & \textbf{0.858} & \textbf{29.51} & \textbf{0.869} & \textbf{0.102} & \textbf{4.610} & 16.20 & 69.27 \\
EVolSplat4D\cite{miao2026evolsplat4defficientvolumebasedgaussian} & 19.71 & 0.603 & 23.24 & 0.715 & 0.288 & 13.870 & 69.65 & 4.78 \\
IGGT\cite{li2025iggt} & 16.09 & 0.273 & 19.33 & 0.560 & 0.442 & 9.615 & 3.40 & 24.50 \\
EPS3D\cite{zhu2026eps3dendtoendfeedforward3d} & 18.46 & 0.533 & 23.60 & 0.725 & 0.213 & 9.812 & 3.48 & 41.81 \\
\bottomrule
\end{tabular}
\endgroup
\end{table*}

\begin{figure*}[t]
\centering
\includegraphics[width=\textwidth]{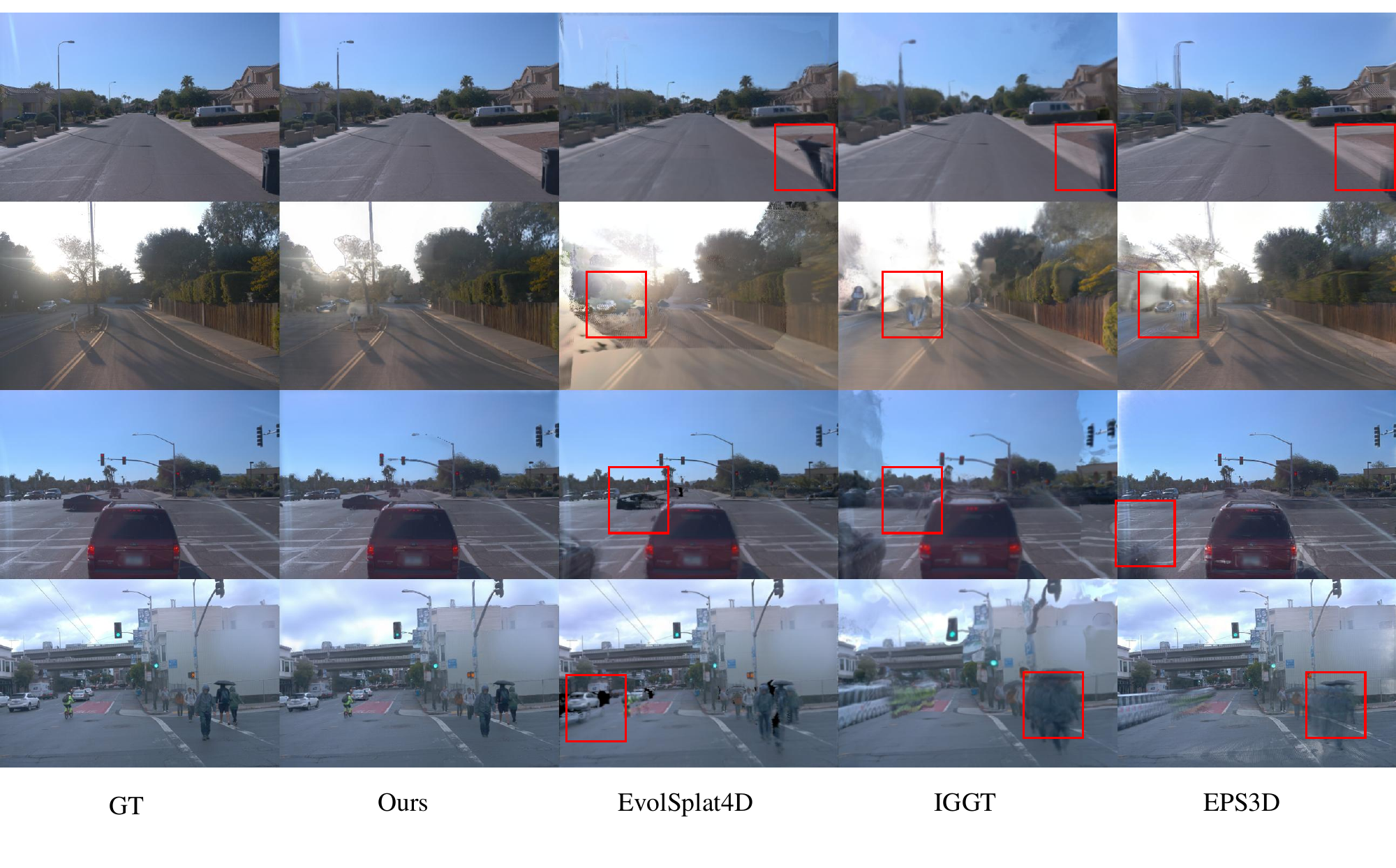}
\caption{\textbf{Qualitative comparison with additional methods on the Waymo Open Dataset.} IGRecon4D better preserves traffic participants and static background details than feed-forward baselines, with fewer participant-region omissions and sharper object boundaries.}
\label{fig:more_comparison}
\end{figure*}

\subsection{More Ablation Study Results}
\label{app:more_ablation}

Table~\ref{tab:more_ablation} and Fig.~\ref{fig:more_ablation} present three additional ablations of the dynamic-mask, geometric-input, and sky-rendering components. In the \emph{Directly using SAM3} variant, we run SAM3~\cite{carion2025sam3segmentconcepts} at inference and take the union of its predicted instance masks as the dynamic mask. The \emph{RGB input only} variant receives RGB images without camera parameters or LiDAR measurements. The \emph{w/o sky gaussians} variant omits sky Gaussians during rendering.

\begin{table*}[t]
\centering
\caption{\textbf{Additional ablation study on the Waymo Open Dataset.} Traffic-participant-region (TPR) metrics are computed on SAM3-generated traffic-participant masks, while full-image metrics are computed over all pixels.}
\label{tab:more_ablation}
\begin{tabular}{lcccccc}
\toprule
Method & TPR PSNR $\uparrow$ & TPR SSIM $\uparrow$ & PSNR $\uparrow$ & SSIM $\uparrow$ & LPIPS $\downarrow$ & D-RMSE $\downarrow$ \\
\midrule
Ours & \textbf{26.76} & \textbf{0.858} & \textbf{29.51} & \textbf{0.869} & \textbf{0.102} & 4.610 \\
Directly using SAM3\cite{carion2025sam3segmentconcepts} & 26.52 & 0.846 & 29.14 & 0.860 & 0.106 & \textbf{4.487} \\
RGB input only & 22.06 & 0.778 & 29.22 & 0.858 & 0.103 & 6.896 \\
w/o sky gaussians & 25.72 & 0.848 & 9.31 & 0.603 & 0.302 & 4.599 \\
\bottomrule
\end{tabular}
\end{table*}

\begin{figure*}[t]
\centering
\includegraphics[width=\textwidth]{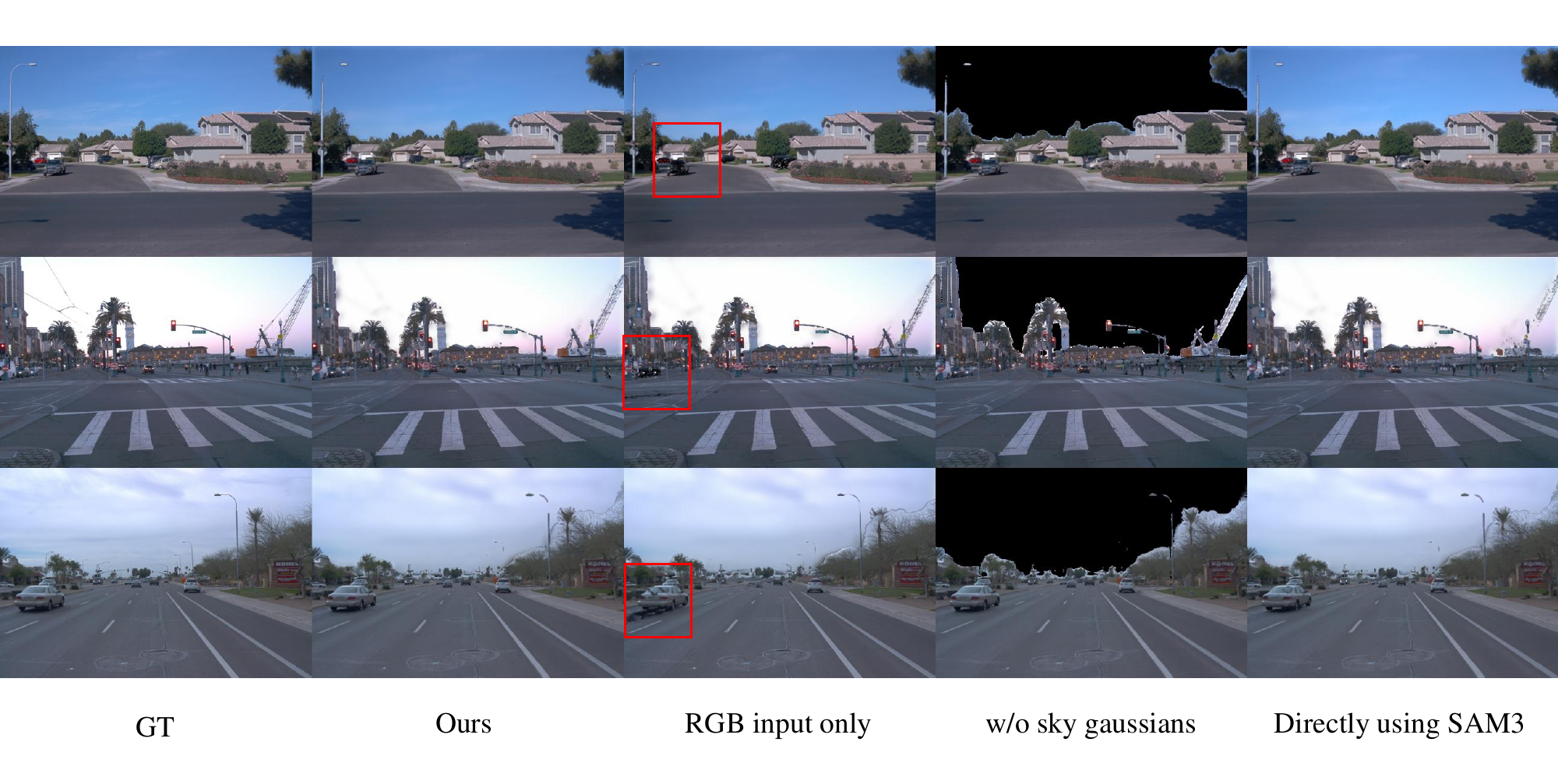}
\caption{\textbf{Additional qualitative ablation results.} The examples compare IGRecon4D with RGB-only input, removal of sky Gaussians, and direct use of SAM3 masks.}
\label{fig:more_ablation}
\end{figure*}

As shown in Table~\ref{tab:more_ablation}, directly using SAM3 yields reconstruction metrics close to those of our learned dynamic-mask predictor. However, calling SAM3 at inference increases the average runtime from 16.20~s for our method to 34.17~s, more than doubling the cost. With RGB input only, TPR PSNR drops to 22.06 and D-RMSE rises to 6.896, showing the value of the camera and LiDAR inputs for reconstruction. Omitting sky Gaussians sharply reduces full-image PSNR to 9.31 while leaving TPR PSNR at 25.72, indicating that the sky representation is particularly important for full-image rendering. Fig.~\ref{fig:more_ablation} provides qualitative examples of these effects.

\subsection{Temporal Interpolation Results}

We evaluate temporal interpolation by reconstructing and rendering a sequence of 21 consecutive frames from the front, front-left, and front-right cameras. For each camera, only frames 0, 5, 10, 15, and 20 are used as reconstruction inputs, while all 21 frames are used for evaluation. At every timestamp, we render the static Gaussians reconstructed from the input frames. At input timestamps, we use the dynamic Gaussians reconstructed from those frames. At intermediate timestamps, following DGGT~\cite{chenfeedforward}, we use pretrained TAPIP3D~\cite{tapip3d} to estimate dynamic-Gaussian trajectories between adjacent input frames and interpolate Gaussian positions and rotations along these trajectories.

Table~\ref{tab:temporal_interp} and Fig.~\ref{fig:temporal_interp} compare our method with DGGT under the same temporal-interpolation setup. Our method achieves better traffic-participant-region and full-image reconstruction metrics, as well as lower D-RMSE, consistent with its stronger dynamic-object reconstruction and geometric consistency.

\begin{table}[!ht]
\centering
\caption{\textbf{Temporal interpolation comparison on the Waymo Open Dataset.} Traffic-participant-region (TPR) metrics are computed on SAM3-generated traffic-participant masks, while full-image metrics are computed over all pixels.}
\label{tab:temporal_interp}
\begin{tabular}{lcc}
\toprule
Metric & Ours & DGGT\cite{chenfeedforward} \\
\midrule
TPR PSNR $\uparrow$ & \textbf{20.57} & 18.95 \\
TPR SSIM $\uparrow$ & \textbf{0.654} & 0.575 \\
PSNR $\uparrow$ & \textbf{25.62} & 21.83 \\
SSIM $\uparrow$ & \textbf{0.774} & 0.699 \\
LPIPS $\downarrow$ & \textbf{0.163} & 0.212 \\
D-RMSE $\downarrow$ & \textbf{5.058} & 6.718 \\
\bottomrule
\end{tabular}
\end{table}

\begin{figure}[!ht]
\centering
\includegraphics[width=\columnwidth]{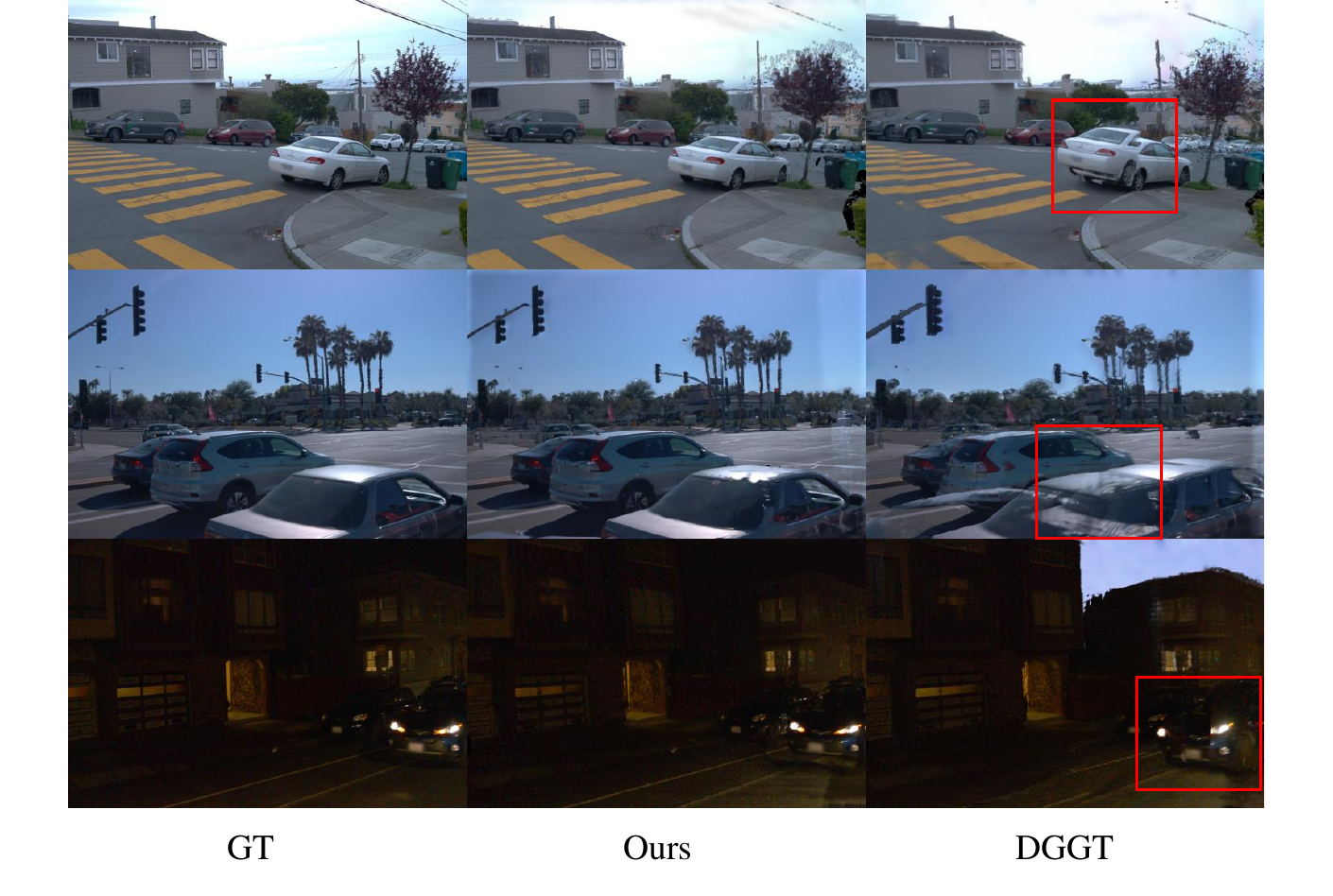}
\caption{\textbf{Qualitative comparison of temporal interpolation results.} Ground truth (GT), IGRecon4D, and DGGT are shown from left to right. Red boxes highlight regions where DGGT produces visible artifacts on traffic participants.}
\label{fig:temporal_interp}
\end{figure}

\subsection{Evaluation of Predicted Dynamic Mask}

We further evaluate the quality of the dynamic masks predicted under IDC supervision. Using pseudo labels generated by SAM3~\cite{carion2025sam3segmentconcepts} as the reference, we compute the accuracy and recall of our predicted masks and compare them with those predicted by the baseline DGGT~\cite{chenfeedforward}. As shown in Table~\ref{tab:dynamic_mask} and Fig.~\ref{fig:dynamic_mask}, our method achieves higher accuracy and recall, indicating more reliable localization of potentially dynamic traffic participants in driving scenes.

\begin{table}[!ht]
\centering
\caption{\textbf{Evaluation of predicted dynamic masks.} Accuracy and recall are measured against SAM3-generated pseudo ground truth; higher is better for both metrics.}
\label{tab:dynamic_mask}
\begin{tabular}{lcc}
\toprule
Method & Accuracy $\uparrow$ & Recall $\uparrow$ \\
\midrule
Ours & \textbf{90.26\%} & \textbf{84.88\%} \\
DGGT\cite{chenfeedforward} & 50.47\% & 34.53\% \\
\bottomrule
\end{tabular}
\end{table}

\begin{figure}[!ht]
\centering
\includegraphics[width=\columnwidth]{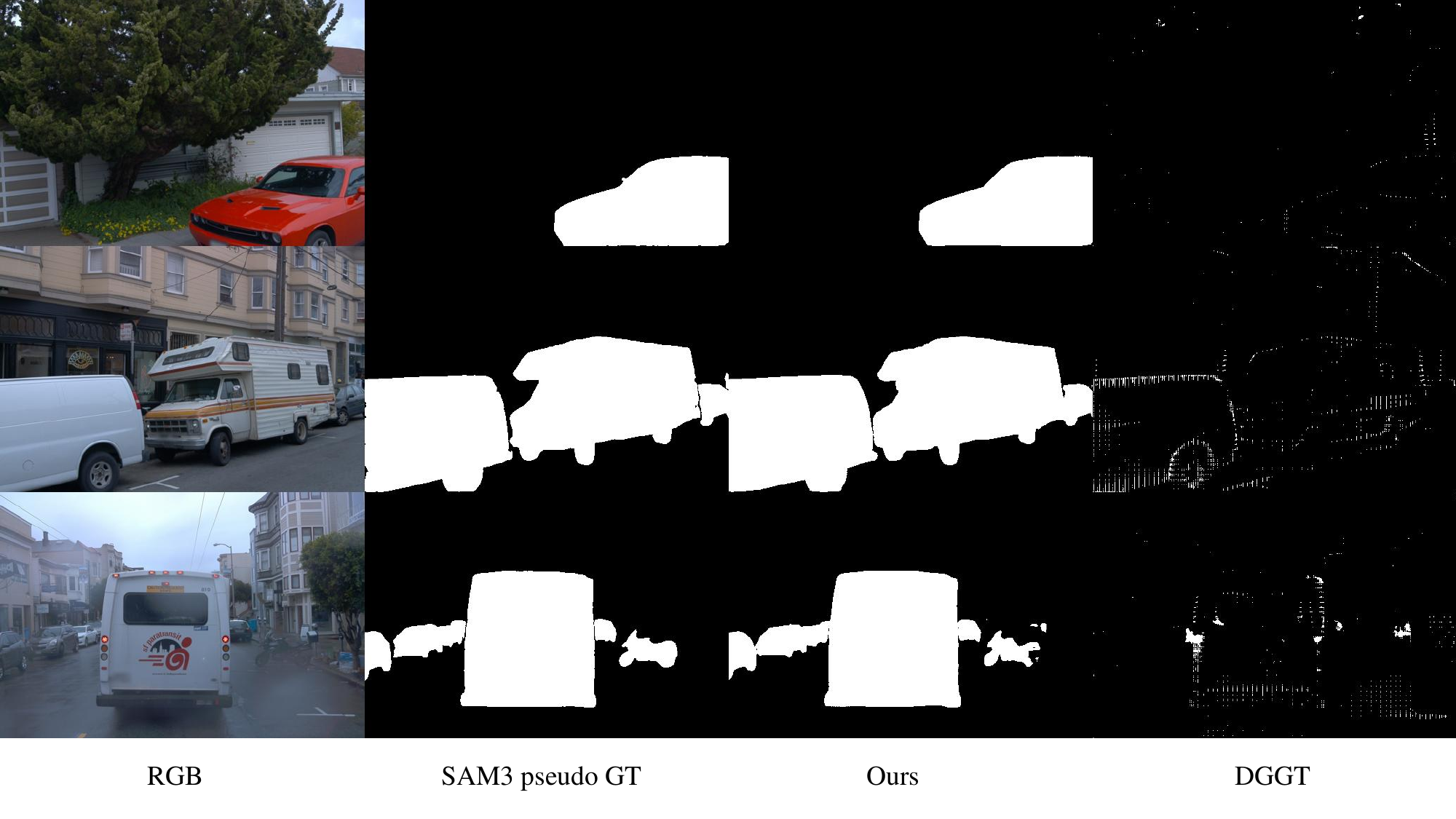}
\caption{\textbf{Qualitative evaluation of predicted dynamic masks.} RGB images, SAM3-generated pseudo ground-truth masks, our predictions, and DGGT predictions are shown from left to right.}
\label{fig:dynamic_mask}
\end{figure}

\subsection{Failure Cases}

Our method still struggles in challenging scenes with strong lighting effects. Fig.~\ref{fig:failure_case} shows a nighttime scene illuminated by streetlights. Although our reconstruction captures the approximate scene geometry, it does not adequately model the lighting. Consequently, the rendered image fails to reproduce the illumination cast by the streetlights, limiting the overall reconstruction quality.

\begin{figure}[!ht]
\centering
\includegraphics[width=\columnwidth]{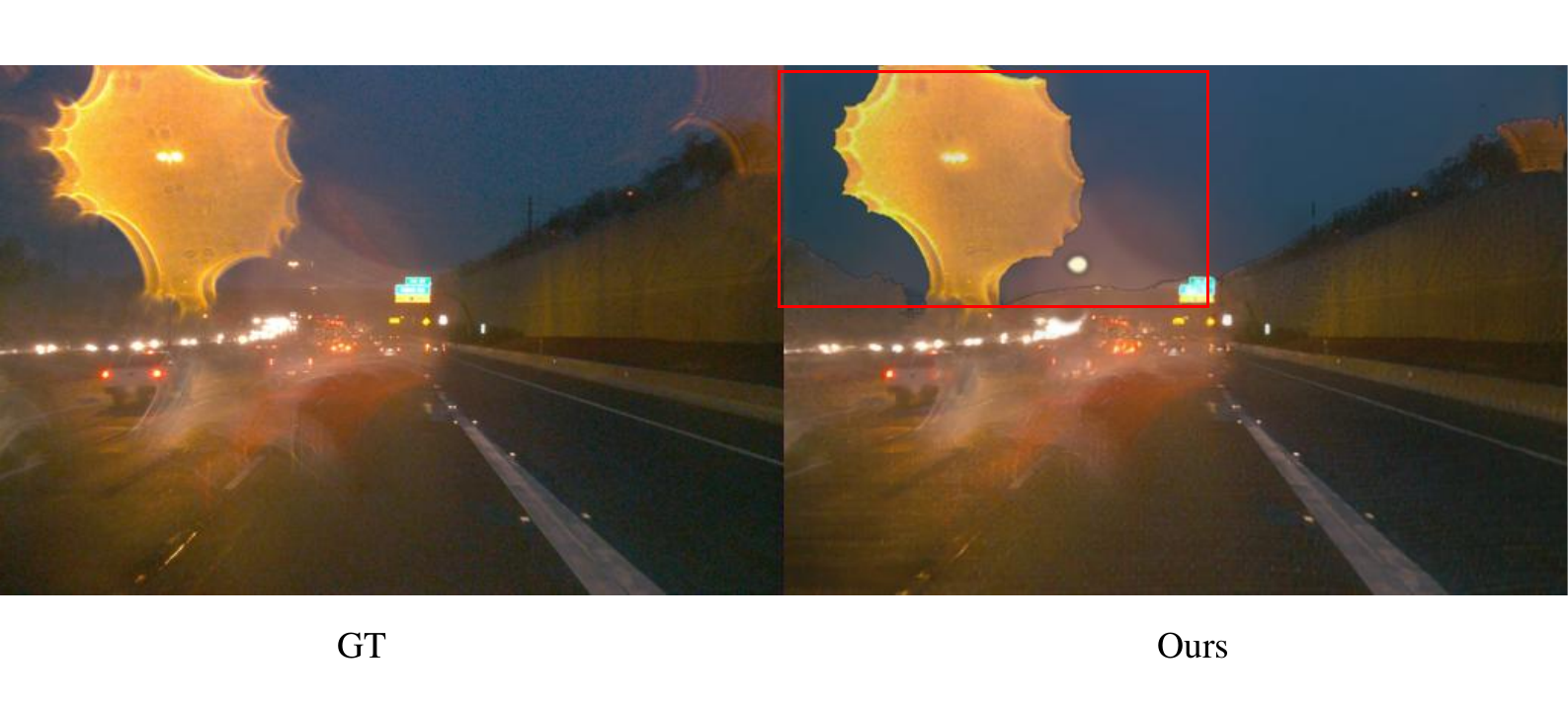}
\caption{\textbf{Failure case in a nighttime scene.} Ground truth (GT) and our reconstruction are shown side by side. The red box highlights a visible reconstruction artifact.}
\label{fig:failure_case}
\end{figure}
\endgroup

\subsection*{Availability of data and materials}

The data that support the findings of this study can be downloaded from their official websites.

\subsection*{Author contributions}

Yifei Tong was responsible for Methodology, Software, Investigation, and Writing. Ziyang Huang contributed to Methodology guidance and Writing. Yizhao Han contributed to Software implementation and Investigation by assisting in training the proposed framework. Xiaoxiao Long and Hao Zhu contributed to Methodology guidance and Writing. All the authors have approved the final manuscript.

\subsection*{Acknowledgements}
The authors would like to thank Nanjing University and Huawei Yinwang for providing experimental resources and support for this research.

\subsection*{Declaration of competing interest}

The authors have no competing interests to declare that are relevant to the content of this article.


\bibliographystyle{CVMbib}
\bibliography{refs}

\subsection*{Author biography}

\makeatletter
\def\@CVMBIOhangdepth{1.25in}
\patchcmd{\biography}
  {\begin{minipage}[b][\@CVMBIOphotodepth][c]{\@CVMBIOphotowidth}}
  {\begin{minipage}[b][\@CVMBIOphotodepth][t]{\@CVMBIOphotowidth}\vspace{0pt}}
  {}{}
\patchcmd{\biography}
  {\begin{minipage}[b][\@CVMBIOphotodepth][c]{\@CVMBIOphotowidth}}
  {\begin{minipage}[b][\@CVMBIOphotodepth][t]{\@CVMBIOphotowidth}\vspace{0pt}}
  {}{}
\makeatother

\begin{biography}[author1]{Yifei Tong} is currently pursuing the M.S. degree in Intelligent Science and Technology at Nanjing University. His research interests include world models and dynamic scene reconstruction.
\end{biography}

\vspace{2.5\baselineskip}

\begin{biography}[coauthor1]{Xiaoxiao Long} is tenure-track associate professor at the School of Intelligence Science and Technology, Nanjing University. He obtained his Ph.D. in 2023 from the Department of Computer Science at the University of Hong Kong, advised by Prof. Wenping Wang (IEEE \& ACM Fellow) and Prof. Taku Komura. He received his bachelor's degree from Zhejiang University in 2018. His long-term research interests include robotics, spatial intelligence, and world models. 
\end{biography}

\vspace{2.5\baselineskip}

\begin{biography}[coauthor2]{Hao Zhu} is an associate professor and Zhicheng Scholar at the School of Intelligence Science and Technology, Nanjing University, where he co-direct the 3D Vision Lab (NJU-3DV) with Prof. Yao Yao and Prof. Long Xiaoxiao. His research interests include computer vision and machine learning, especially for 3D reconstruction and 3D vision.
\end{biography}

\end{document}